%% file: main.tex
\documentclass[journal]{IEEEtran} 
\usepackage[noadjust]{cite}

\usepackage[section]{placeins}
\usepackage{float}
\usepackage{flushend}
\ifCLASSINFOpdf
  \usepackage[pdftex]{graphicx}
  \graphicspath{{./fig/}{./result/}}
\else
\fi
\usepackage{siunitx} 
\usepackage{algorithmic}

\usepackage{array}

\ifCLASSOPTIONcompsoc
  \usepackage[caption=false,font=normalsize,labelfont=sf,textfont=sf]{subfig}
\else
  \usepackage[caption=false,font=footnotesize]{subfig}
\fi
\usepackage[utf8]{inputenc} 
\usepackage[T1]{fontenc}    
\usepackage{booktabs}       
\usepackage{amsfonts}       
\usepackage{nicefrac}       
\usepackage{microtype}      
\usepackage{subfloat}      %
\usepackage{mathtools}
\usepackage{amsmath}
\usepackage{multicol}

\usepackage{cancel}
\usepackage[normalem]{ulem}
\usepackage{amsmath}

\usepackage{xcolor}
\usepackage[export]{adjustbox} 
\definecolor{mygreen}{rgb}{0.0,0.5,0.6}
\definecolor{mygray}{rgb}{0.5,0.5,0.5}
\definecolor{mymauve}{rgb}{0.58,0,0.82}

\definecolor{cb_orange}{rgb}{1.0,0.51,0.0}
\definecolor{cb_blue}{rgb}{0,0,1}
\definecolor{cb_green}{rgb}{0.3,0.67,0.29}
\definecolor{cb_red}{rgb}{0.89,0.1,0.11}
\definecolor{cb_pink}{rgb}{1, 0, 0.4}
\definecolor{cb_gray}{rgb}{0.8, 0.8, 0.8}

\newif\ifclean

\cleantrue 

\ifclean
\newcommand{\stkout}[1]{}
\newcommand{\del}[1]{}
\newcommand{\rv}[1]{#1} 
\else
\newcommand{\stkout}[1]{\ifmmode\text{\sout{\ensuremath{#1}}}\else\sout{#1}\fi}
\newcommand{\del}[1]{\stkout{#1}}
\newcommand{\rv}[1]{\textcolor{cb_blue}{#1}}
\fi

\newif\ifcleant
\cleanttrue 

\ifcleant
\newcommand{\stkoutt}[1]{}
\newcommand{\delt}[1]{}
\newcommand{\rvt}[1]{#1} 
\else
\newcommand{\stkoutt}[1]{\ifmmode\text{\sout{\ensuremath{#1}}}\else\sout{#1}\fi}
\newcommand{\delt}[1]{\stkoutt{#1}}
\newcommand{\rvt}[1]{\textcolor{cb_blue}{#1}}
\fi
\usepackage{xspace}
\newcommand{\placeholder}{RefineGAN\xspace}

\begin{document}
\bstctlcite{IEEEexample:BSTcontrol}
%
\title{Compressed Sensing MRI Reconstruction using a Generative Adversarial Network with
a Cyclic Loss}
%
%
%


\author{Tran~Minh~Quan,~\IEEEmembership{Student Member,~IEEE,}
	    Thanh~Nguyen-Duc, 
	    and~Won-Ki~Jeong,~\IEEEmembership{Member,~IEEE}

\thanks{Tran Minh Quan, Thanh Nguyen-Duc and Won-Ki Jeong are with Ulsan National Institute of Science and Technology (UNIST).}
\thanks{E-mail: \{quantm, thanhnguyencsbk, wkjeong\}@unist.ac.kr}
}
%
%

{}
%



\maketitle

\begin{abstract}
\input{abstract}

\end{abstract}


%
\IEEEpeerreviewmaketitle

\renewcommand{\topfraction}{.75}
\renewcommand{\bottomfraction }{.25}

\section{Introduction}
\input{intro}
\section{Related work}
\input{related}

\section{Method}
\input{method}

\section{Result}
\input{result}

\section{Conclusion}
\input{conclusion}
\rvt{
\section*{Acknowledgment}
This research was partially supported by the 2017 Research Fund (1.170017.01) of UNIST, 
the Bio \& Medical Technology Development Program of the National Research Foundation of Korea (NRF) funded by the Ministry of Science and ICT (MSIT) (NRF-2015M3A9A7029725), 
the Next-Generation Information Computing Development Program through the NRF funded by the MSIT (NRF-2016M3C4A7952635), 
and the Basic Science Research Program through the NRF funded by the Ministry of Education (MOE) (NRF-2017R1D1A1A09000841). 
The authors would like to thank Dr. Yoonho Nam for the helpful discussion and MRI data, and Yuxin Wu for the help on Tensorpack. 
}





\bibliographystyle{IEEEtran}
\bibliography{reference}
\vfill

\end{document}

%% file: abstract.tex
%
Compressed Sensing MRI (CS-MRI) has provided theoretical foundations upon which the time-consuming MRI acquisition process can be accelerated. 
However, it primarily relies on iterative numerical solvers which still hinders their adaptation in time-critical applications.
In addition, recent advances in deep neural networks have shown their potential in computer vision and image processing, but their adaptation to MRI reconstruction is still in an early stage.
In this paper, we propose a novel deep learning-based generative adversarial model, \textit{\placeholder}, for fast and accurate CS-MRI reconstruction. 
The proposed model is a variant of fully-residual convolutional autoencoder and generative adversarial networks (GANs), specifically designed for CS-MRI formulation; 
it employs deeper generator and discriminator networks with cyclic data consistency loss for faithful interpolation in the given under-sampled $k$-space data.
In addition, our solution leverages a chained network to further enhance the reconstruction quality. 
\textit{\placeholder} is fast and accurate -- the reconstruction process is extremely rapid, as low as tens of milliseconds for reconstruction of a 256x256 image, because it is one-way deployment on a feed-forward network, and the image quality is superior even for extremely low sampling rate (as low as 10\%) due to the data-driven nature of the method.
%
We demonstrate that \textit{\placeholder} outperforms the state-of-the-art CS-MRI methods by a large margin in terms of both running time and image quality via evaluation using several open-source MRI databases. 

%% file: intro.tex

%
\IEEEPARstart{M}{agnetic} resonance imaging (MRI) has been widely used as an in-vivo imaging technique because it is non-intrusive, high-resolution, and safe to living organisms.
%
Even though MRI does not use dangerous radiation for imaging, its long acquisition time causes discomfort to patients and hinders applications in time-critical diagnoses, such as strokes.
%
%
%
To speed up acquisition time, various acceleration techniques have been developed. 
%
One approach is using parallel imaging hardware~\cite{Heidemann2003} 
to reduce time-consuming phase-encoding steps. 
Another approach is adopting  the Compressive Sensing (CS) theory~\cite{donoho2006compressed} 
to MRI reconstruction~\cite{lustig2008compressed} so that only a small fraction of data is needed to generate full reconstruction via a computational method. 
Even combining parallel imaging and CS-MRI is studied to maximize the acceleration of acquisition~\cite{MRM:MRM22161}. 
%
%
%
To apply the CS theory to MRI reconstruction, we must find a proper sparsifying transformation to make the signal sparse, e.g., wavelet transformation, and solve an  $\ell_{1}$ minimization problem with regularizers. 

Early work on CS-MRI primarily focused on applying pre-defined universal sparsifying transforms, such as the discrete Fourier transform (DFT), discrete cosine transform (DCT), total variation (TV), or discrete wavelet transform (DWT)~\cite{daubechies_ten_1992}, and developing efficient numerical algorithms to solve nonlinear optimization problems~\cite{goldstein2009split, boyd2011distributed}.
%
More recently, data-driven sparsifying transforms (i.e., dictionary learning) 
have gained much attention in CS-MRI due to their ability to express local features of reconstructed images more accurately compared to pre-defined universal transforms.
%
In dictionary learning based CS-MRI, the reconstructed images are approximated using either patch-based atoms or convolution filters, and the results are generated by training the dictionary jointly (blindly) with the reconstructed images or by using a pre-trained set of atoms from the database.
%
%
Even though dictionary learning based methods show much improved image quality, the reconstruction process still suffers from longer running time due to the extra computational overhead for dictionary training and sparse coding. 

%


%
\stkoutt{
In the last five years, deep learning~\cite{lecun_deep_2015} has gained substantial attention, primarily because it has surpassed human performance in solving many complex problems. 
Deep learning is accomplished by forming a deep neural network from many perceptron layers and then training the model for a particular task.
In visual recognition tasks, this architecture is able to learn hierarchically to extract patterns such as handwritten digits and other features of interest~\cite{krizhevsky_imagenet_2012} from images~\cite{zeiler_visualizing_2014}. 
The main drawback of deep neural networks is that they require an enormous amount of data for adequate training.
To overcome this issue, researchers have begun to collect large databases~\cite{russakovsky_imagenet_2015} containing millions of images from hundreds of categories.
These rich databases have enabled several advanced architectures, including VGG~\cite{vgg_simonyan_2014}, GoogLeNet~\cite{szegedy_going_2015}, ResNet~\cite{he_residual_2015}, and many more.
With such improvements, computers became able to mimic artistic painting styles by transferring complicated properties from one image to another~\cite{gatys_image_2016}. 
In addition, researchers have extended deep learning methods to medical image analysis~\cite{cicek_3d_2015}, enabling automatic classification and segmentation of data from modalities such as computed tomography (CT)~\cite{zheng_3d_2015} and MRI~\cite{isin_review_2016}. 
}

%
\rvt{
The primary motivation for the proposed work stems from the following observations: recent advances in deep learning~\cite{lecun_deep_2015} have yielded encouraging results for many computer vision problems, which shows a potential to ``shift'' the time-consuming computing process into the training (pre-processing) phase and to reduce prediction time by performing only one-pass deployment 
of neural network instead of using iterative methods commonly used in conventional image processing methods.
}
In addition, the past success of deep learning in single-image super resolution, denoising, and in-painting are in line with the analogy of CS-MRI reconstruction, which is about prediction of missing information from the incomplete (corrupted) image. 
%
Especially, we discovered that the recently introduced Cycle-Consistent Adversarial Networks (CycleGAN~\cite{zhu2017unpaired}) map naturally to CS-MRI problems to enforce consistency in measurement and reconstruction. 

Based on these observations, in this paper, we propose a novel GAN-based deep architecture for CS-MRI reconstruction that is fast and accurate.
The proposed method builds upon several state-of-the-art deep neural networks, such as convolutional autoencoder, residual  networks, and generative adversarial networks (GANs), with novel \textit{cyclic loss} for data consistency constraints that promotes accurate interpolation of the given undersampled $k$-space data. 
Our proposed network architecture is fully residual, where inter- and intra-layers are linked via addition-based skip connections to learn residuals, so the network depth can be effectively increased without suffering from the gradient vanishing problem and have more expressive power. 
In addition, our generator consists of multiple end-to-end networks chained together where the first network translates a zero-filling reconstruction image to a full reconstruction image, and the following networks improve accuracy of full reconstruction image (i.e., \textit{refining} the result).
%
To the best of our knowledge, the proposed work is the first CS-MRI method employing a cyclic loss with fully residual convolutional GANs that achieved real-time performance (reconstruction of a 256$\times$256 image can be done under 100 ms) with superior image quality (over 42 dB in average for the 40\% sampling rate), which we believe it has a huge potential for time-critical applications.
We demonstrate that our method outperforms recent CS-MRI methods in terms of both running time and image quality via performance assessment on several open-source MRI databases.

%
%

%
The rest of this paper is organized as follows. 
In Section~\ref{sec:related}, we review the recent work related to CS-MRI algorithms, with and without support from deep learning. 
%
Section~\ref{sec:method} introduces our proposed method in detail.
%
Finally, we show the results and compare the performance of our method with the performance of other methods in  Section~\ref{sec:result}. 
We summarize our work and suggest future research directions in Section~\ref{sec:conclusion}. 

%% file: related.tex

\label{sec:related}

The current CS-MRI methods 
can be broadly classified into three categories: conventional $\ell_{1}$ energy minimization approach using universal sparsifying transformation, machine learning-based approach using the dictionary and sparse coding, and deep learning-based approach using state-of-the-art deep neural networks.

\textbf{Universal transform-based methods:}
%
%
The long acquisition time is a fundamental challenge in MRI, so the compressed sensing (CS) theory has been proposed and successfully applied to speed up the acquisition process.
Conventional CS-MRI reconstruction methods have been developed to leverage the sparsity of signal by using universal sparsifying transforms, such as Fourier transform, Total Variation (TV), and Wavelets~\cite{lustig2008compressed}, and to exploit the spatio-temporal correlations, such as $k-t$ FOCUSS~\cite{jung2007improved,jung2009ktgeneral}
This sparsity-based CS-MRI method introduces computational overhead in the reconstruction process due to solving expensive nonlinear $\ell_{1}$ minimization problem, which leads to developing efficient numerical algorithms~\cite{goldstein2009split} and adopting parallel computing hardware to accelerate the computational time~\cite{quan2015multi}.
The nuclear norm and low-rank matrix completion techniques have been employed for CS-MRI reconstruction as well~\cite{yao2015accelerated, otazo2015low, tremoulheac2014dynamic} .  
%

%

\textbf{Dictionary learning-based methods:}
The main limitation of universal transform-based methods is that the transformation is general and not specifically designed for the input data. 
In contrast, dictionary learning~\cite{aharon2006ksvd} (DL) can generate data-specific dictionary and improve the image quality. 
Earlier work using DL in CS-MRI is using image patches to train dictionary~\cite{awate2012spatiotemporal, caballero2014dictionary, ravishankar2011mr}, which may suffer from redundant atoms and longer running time. 
More recently, convolutional sparse coding  (CSC), a new learning-based sparse representation, approximates the input signal with a superposition of sparse feature maps convolved with a collection of filters.
CSC is shift-invariant and can be efficiently computed using parallel algorithms in the frequency domain. 
2D and 3D CSC have been successfully adopted in dynamic CS-MRI~\cite{quanisbi, quan2016miccai}, and efficient numerical algorithms, such as the alternating direction method of multipliers, are proposed to further accelerate CSC computation in CS-MRI~\cite{bristow2013fast, wohlberg2014efficient}


\textbf{Deep learning-based methods:}
Deep learning-based CS-MRI is aimed to design fast and accurate method that reconstruct high-quality MR images from under-sampled $k$-space data using multi-layer neural networks. 
Earlier work using deep learning in CS-MRI is mostly about the direct mapping between a zero-filling reconstruction image to a full-reconstruction image using a deep convolutional autoencoder network~\cite{wang2016accelerating}. 
Lee \textit{et al.}~\cite{lee2017deep} proposed a similar autoencoder-based model but the method learns noise (i.e., residual) from a zero-filling reconstruction image to remove undersampling artifacts.
Another interesting deep learning-based CS-MRI approach is Deep ADMM-Net~\cite{sun2016deep}, which is a deep neural network architecture that learns parameters of the ADMM algorithm (e.g., penalty parameters, shrinkage functions, etc.) by using training data. 
This deep model consists of multiple stages, each of which corresponds to a single iteration of the ADMM algorithm. 
%
There exist more deep learning-based MR reconstruction work; for example, Schlemper \textit{et al.}~\cite{schlempler2017deepcascade} proposed a model that cascades a multiple deep network with data consistency layer, Lee \textit{et al.}~\cite{lee2017parallel} proposed learning the artifact pattern instead of aliasing-free MR image, and Jin \textit{et al.}~\cite{jin2017fbpconv} used deep learning to model the inverse problems in medical imaging. 
%
Recently, Generative Adversarial Nets~\cite{goodfellow2014generative} (GANs), a general framework for estimating generative models via an adversarial process, has shown outstanding performance in image-to-image translation.
Unsupervised variants of GANs, such as DiscoGAN~\cite{kim2017learning} and CycleGAN~\cite{zhu2017unpaired}, have been proposed for mapping different domains without matching data pairs.
Inspired by their success in image processing, GANs have been employed for reconstructing zero-filling under-sampled MRI with~\cite{mardani2017deep} and without~\cite{yu2017deep} the consideration of data consistency during the training process.  
As shown above, deep learning has proven itself very promising in CS-MRI as well for reducing reconstruction time while maintaining superior image quality. However, its adaptation in CS-MRI is still in its early stage, which leaves room for improvement.

%% file: method.tex

\label{sec:method}

%
%
\begin{figure*}[t]
	\centering
	\includegraphics[width=1\textwidth]{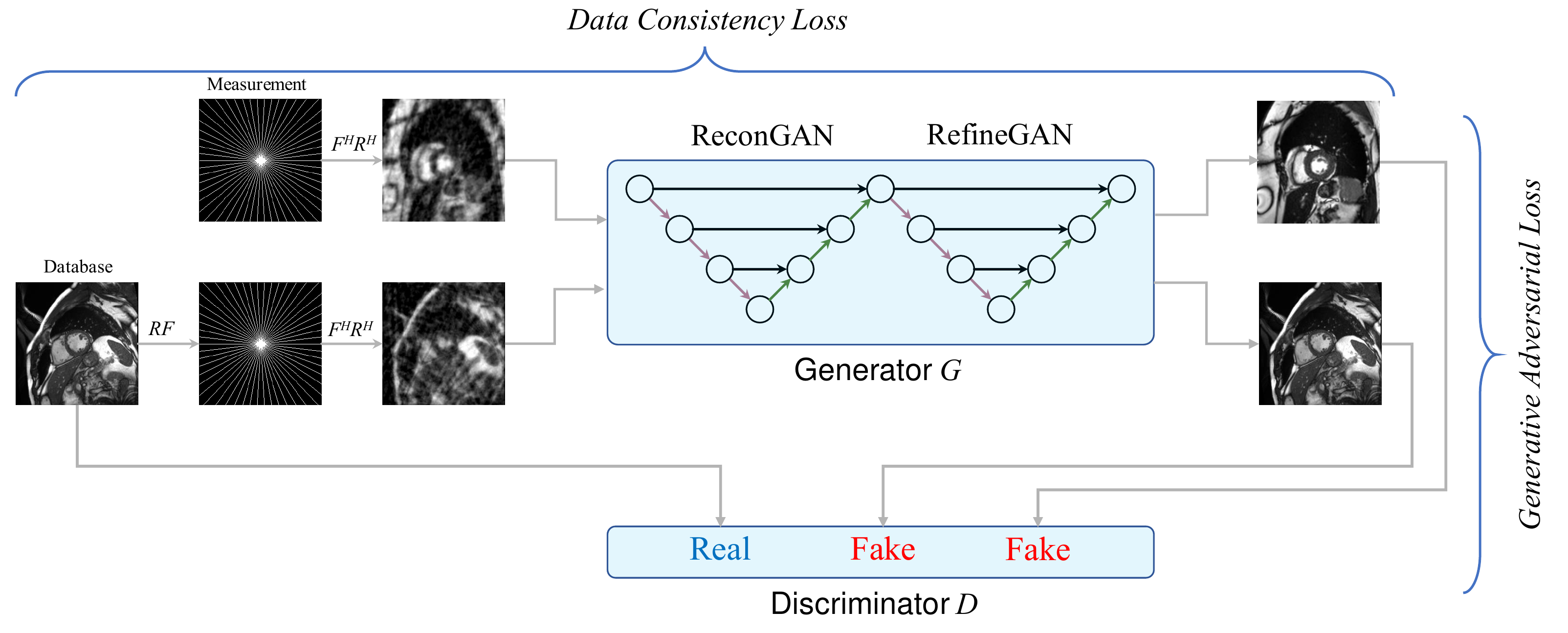}
	\caption{Overview of the proposed method: it aims to reconstruct the images which are satisfied the constraint of under-sampled measurement data; and whether those look similar to the fully aliasing-free results. Additionally, if the fully sampled images taken from the database go through the same process of under-sampling acceleration; we can still receive the reconstruction as expected to the original images.}
	\label{fig:overview}	
\end{figure*}

\subsection{Problem Definition and Notations}
\label{sec:background}

\stkout{
In this section, we briefly overview the basics of Compressed Sensing MRI and notations.
In the following discussion, we use the subscript $f$ to denote the result of applying a masked Fourier transform to a given variable.}
\rv{We denote the under-sampled raw MRI data ($k$-space measurements) using the sampling mask $R$ as $m$. }
\rv{Then, its zero-filling reconstruction $s_0$ can be obtained by the following equation:
\begin{equation}
s_0 = {F^H}{R^H}\left( {{m}} \right)
\end{equation}
where $F$ is the Fourier operator, and superscript $H$ indicates the conjugated transpose of a given operator. 
%
Similarly, turning any image $s_i$ into its under-sampled measurement $m_{s_i}$ with the given sampling mask $R$ can be done via the inverse of the reconstruction process: 
\begin{equation}
{m_{s_i}} = RF\left( s_i \right)
\end{equation}
}
%
%
%
%
\rv{
Then, compressed sensing MRI reconstruction, which is a process of generating a full-reconstruction image $s$ from under-sampled $k$-space data $m$, can be described as follows:
\begin{equation}
\mathop {\min }\limits_s  {J}\left( {s} \right)\quad s.t.\quad {RF(s)} = {m}\
\end{equation}
}
\stkout{where $F$ is the Fourier operator, $R$ is the sampling mask, and superscript $H$ indicates the conjugated transpose of a given operator.
and the above constrained problem can be reformulated in an unconstrained fashion with weighting parameters as follows:
}
\ifclean
\else
\[
\stkout{
\mathop {\min }\limits_s \left\| {{RF(s)} - {m}} \right\|_2^2 + \lambda J\left( s \right)\quad
}
\]
\fi
\rv{
where ${J\left( s \right)}$ is a regularizer required for ill-posed optimization problems.
In our method, this energy minimization process is replaced by the training process of the neural network.}

%
%
\stkout{ 
Many classical priors can be used for ${J\left( s \right)}$, such as Tikhonov regularization (IID Gaussian prior)~\cite{chaari2008autocalibrated}, edge-preserving regularization~\cite{tian2011low}, total variation (TV)~\cite{chen2013calibrationless}, non-local mean filter (NLM)~\cite{manjon2008mri}, wavelets~\cite{pizurica2006review}, curvelets~\cite{bhadauria2013medical}, etc. 
By enforcing $\ell_p$ norm where $0\leq p \leq 1$ for regularization, compressed sensing theory~\cite{donoho2006compressed, candes2008introduction} can be applied to MRI reconstruction from highly undersampled $k$-space data~\cite{lustig2008compressed}. 
This sparsity-induced image reconstruction method aims to find the solution (images $s$) that satisfies not only the under-sampled measurement constraints but also sparsity in the transformed domain by decomposing the signals with the designated universal transform sparsity basis. 
For example, the seminal work by Lustig \textit{et al.}~\cite{MRM:MRM21391} proposed that ${J\left( s \right)}$ is equal to $ {\left\| {\Psi s} \right\|_0}$, in which $\Psi$ is the wavelet transform, and further relaxed the $\ell_0$ norm by $\ell_1$ norm substitution. 
}

\subsection{Overview of the Proposed Method}
Figure~\ref{fig:overview} represents an overview of the proposed method: \rv{Our generator $G$ consists of two-fold chained networks that generate the full MR image directly from a zero-filling reconstruction image (i.e., image generated from under-sampled $k$-space data), in which each input can be up to 2-channel to represent real- and imaginary image of the complex-valued MRI data.}
The generated result is favorable to the fully-sampled data taken from an extensive database and put through the same under-sampling process. 
In contrast, the discriminator $D$ attempts to differentiate between the real MRI instances from the database and the fake results output generated by $G$. 
The entire system involves training $G$ and $D$ adversarially until a balance is reached at the convergence stage. 
Details of each component will be discussed shortly.

\begin{figure*}
	\centering
	\includegraphics[width=0.8\textwidth]{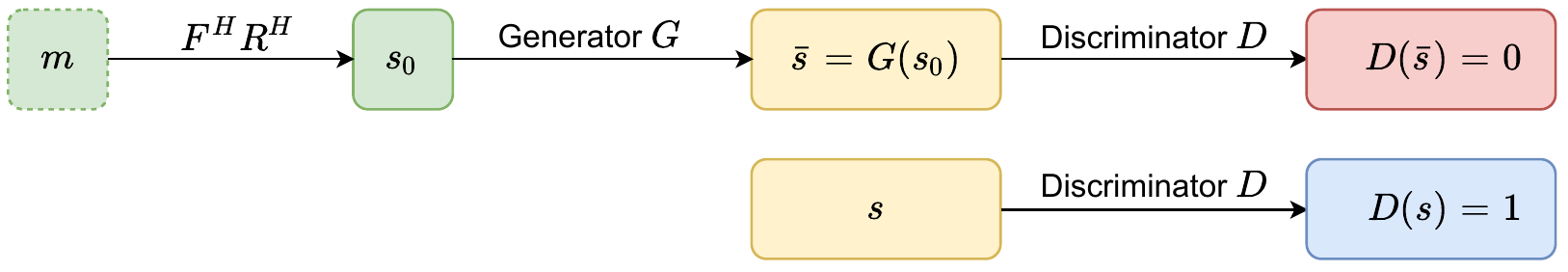}
	\caption{Two learning processes are trained adversarially to achieve better reconstruction from generator $G$ and to fool the ability of recognizing the real or fake MR image from discriminator $D$.}
	\label{fig:ReconGAN}	
\end{figure*}

\begin{figure*}
	\centering
	\includegraphics[width=0.8\textwidth]{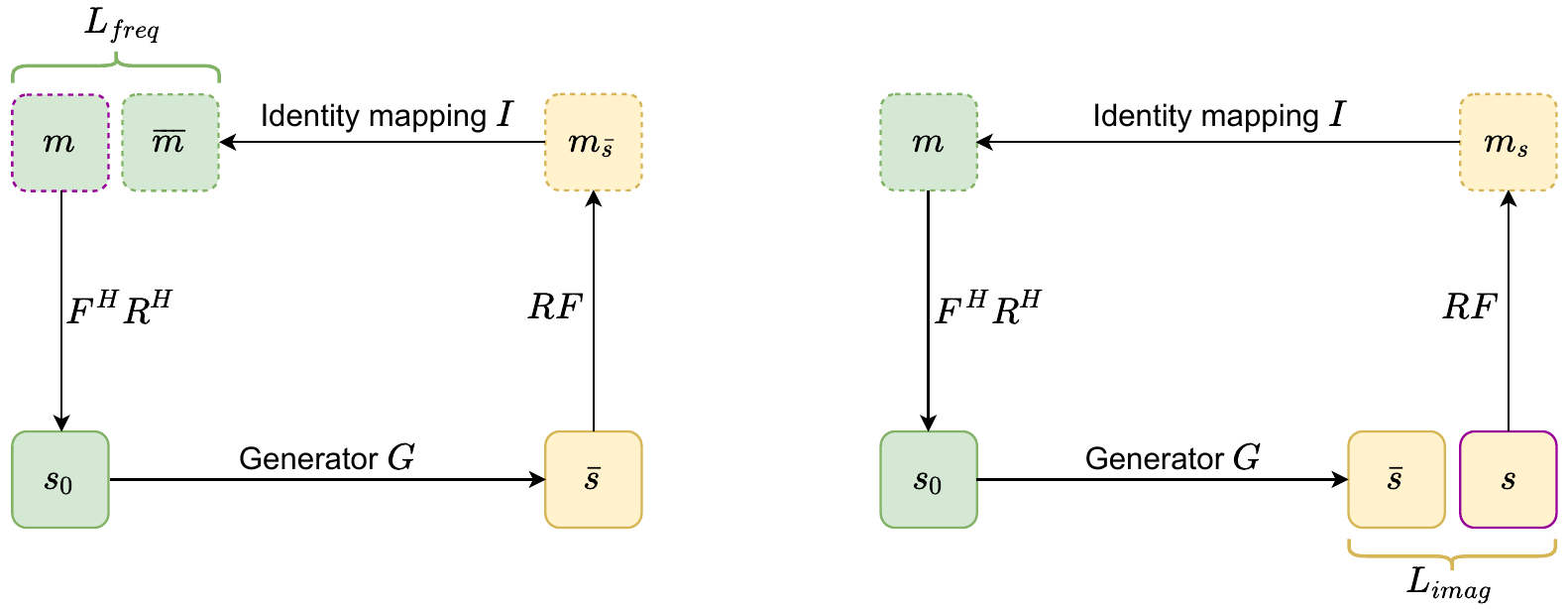}
	\caption{The cyclic data consistency loss, which is a combination of under-sampled frequency loss and the fully reconstructed image loss.}
	\label{fig:dualloss}
\end{figure*}

\subsection{Generative Adversarial Loss}
Our objective is to train generator $G$, which can transform any zero-filling reconstruction $s_0 = F^HR^H(m),~m \in {M}$, where $M$ is the collection of under-sampled $k$-space data, to a fully-reconstructed image $\overline s$ under the constraint that $\overline s$ is \textit{indistinguishable} from all images $s \in S$ reconstructed from full $k$-space data. 
To accomplish this aim, a discriminator $D$ is attached to distinguish whether the image is synthetically generated from $s_0$ by $G$ ($\overline s$, which is considered \textit{fake}) or is reconstructed from fully-sampled $k$-space data ($s$, which is considered \textit{real}).
\rv{
In other words, at each epoch, $G$ tries to produce a reconstruction that can fool $D$ whereas $D$ avoids to be fooled. 
This kind of training borrows the win-lose strategy which is very common in game theory.
}
We wish to train $D$ so that it can maximize the probability of assigning the correct \textit{true} or \textit{false} label to images. 
Note that the objective function for $D$ can be interpreted as maximizing the log-likelihood for estimating
the conditional probability, where the image comes from: $D\left( {\overline s } \right) = D\left( {G\left( s_0 \right)} \right) = 0$ (\textit{fake}), and $D\left( s \right) = 1$ (\textit{real}). 
Simultaneously, generator $G$ is trained to minimize $\left[ {1 - \log D\left( {\overline s } \right)} \right]$ or $\left[ {1 - \log D\left( {G\left( s_0 \right)} \right)} \right]$. 
This can be addressed by formally defining an adversarial loss ${L_{adv}}$, for which we wish to find the solutions of its minimax problem:
\begin{equation}
	\mathop {\min }\limits_G \mathop {\max }\limits_D {L_{adv}}\left( {G,D} \right)
\end{equation}
where
\begin{equation}
\begin{aligned}
{L_{adv}}\left( {G,D} \right) = \mathop E\limits_{m \in {M}} \left[ {1 - \log D\left( {G\left( s_0 \right)} \right)} \right] + \mathop E\limits_{s \in S} \left[ {\log D\left( s \right)} \right]
\label{eq:advloss}
\end{aligned}
\end{equation}

Figure~\ref{fig:ReconGAN} is a schematic depiction of our adversarial process: $G$ tries to generate images $\overline s  = G\left( s_0 \right)$ look similar to the images $s$ that have been reconstructed from full $k$-space data, while $D$ aims to distinguish between $\overline s$ and $s$. 
%
Once the training converges, $G$ can produce the result $\overline s$ that is close to $s$, and $D$ is unable to differentiate between them, which results in bringing the probability for both real and fake labels to 50\%. 
\rv{
In practice, Wasserstein GAN~\cite{arjovsky2017wgan} energy is commonly used to improved the stability of the training processes and is also adopted to our method.}

\subsection{Cyclic Data Consistency Loss}
In an extreme case, with large enough resources and data, the network can map the zero-filling reconstruction $s_0$ 
to any existing fully reconstructed images $s \in S$. 
Therefore, the adversarial loss alone is not sufficient to correctly map the under-sampled data $s_0[n]$ and the full reconstruction $\overline s[n]$ for all $n$. 
To strengthen the bridge connection between $s_0[n]$ and $\overline s[n]$, we introduce an additional constraint, the data consistency loss $L_{cyc}$, which is a combination of under-sampled frequency loss $L_{freq}$ and fully reconstructed image loss $L_{imag}$ in a cyclic fashion.
The first term $L_{freq}$ guarantees that when we perform another under-sampled operator $RF$ on reconstructed images $\overline s[i]$ to get $\overline m[i]$, the difference between $\overline m[i]$ and $m[i]$ should be minimal. 
The second energy term $L_{imag}$ ensures that for any other images $s[j] \in S$ taken from the fully reconstructed data, if $s[j]$ goes through the under-sampling process (by applying $RF$), and the generator $G$ takes zero-filling reconstruction $s_0[j]$ (by applying $F^HR^H$ to $m[j]$) to produce the reconstruction $\overline s[j]$, then both $\overline s[j]$ and $s[j]$ should appear to be similar. 
Those two losses are described in a cyclic fashion in Figure~\ref{fig:dualloss}. 
In practice, various distance metrics, such as mean-square-error (MSE), mean-absolute-error (MAE), etc, can be employed to implement $L_{cyc}$:
\begin{equation}
\begin{aligned}
{L_{cyc}}\left( G \right) &= {L_{freq}}\left( G \right) + {L_{imag}}\left( G \right) \\
&= {\bf{d}} \left( {{m}\left[ i \right],{{\overline m }}\left[ i \right]} \right) +  {\bf{d}} \left( {s\left[ j \right],\overline s \left[ j \right]} \right)
\label{eq:}
\end{aligned}
\end{equation}
Note that the data consistency loss only affects the generator $G$ and not the discriminator $D$. 
Furthermore, each individual loss $L_{freq}$ or $L_{imag}$ is evaluated on its own samples: $m[i]$ and $s[j]$ are drawn independently from $M$ and $S$. 
\rv{Since they are non-complex pixel-wise distance metrics, they will return scalars regardless the input data are either magnitude- or complex-valued numbers.}

\begin{figure*}[t]
	\centering
	\includegraphics[width=1\linewidth]{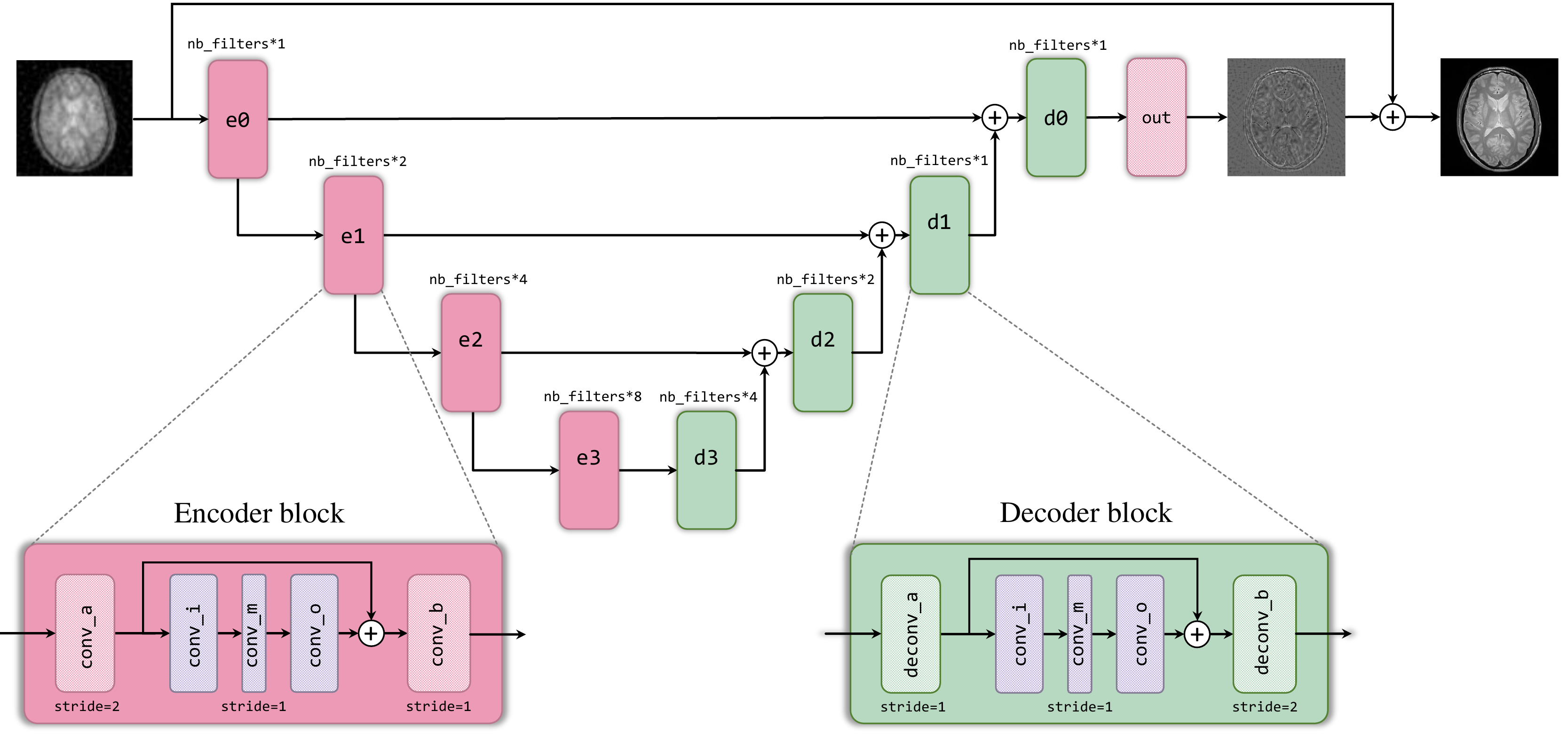}
	\caption{Generator $G$, built by basic building blocks, can reconstruct inverse amplitude of the 
    \rv{residual component} causes by reconstruction from under-sampled $k$-space data. The final result is obtained by adding the zero-filling reconstruction to the output of $G$}
	\label{fig:G}	
\end{figure*}


%
%
%

\begin{figure*}[t]
	\centering
	\subfloat[ReconGAN]
	{
		\includegraphics[width=0.36\linewidth]{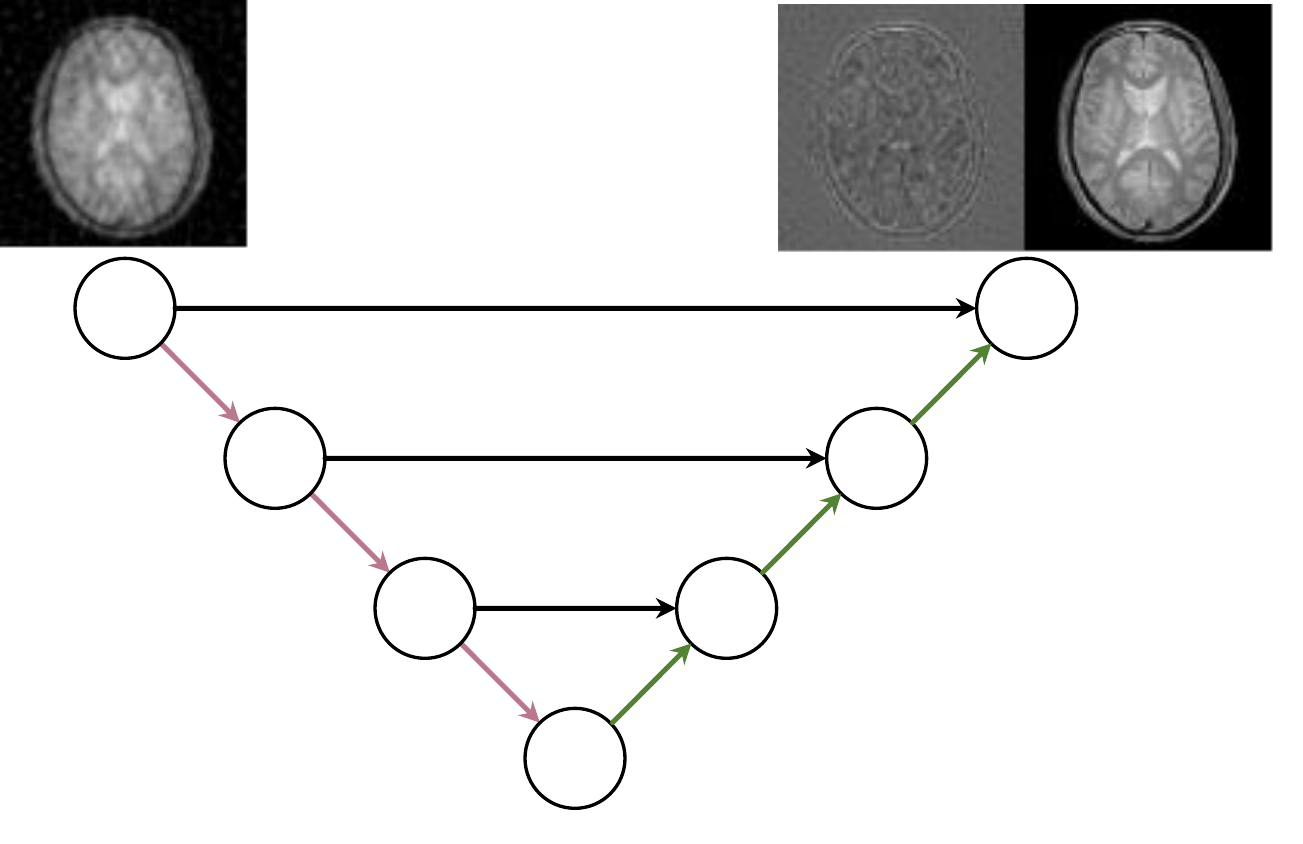}
	}
	\subfloat[RefineGAN]
	{
		\includegraphics[width=0.6\linewidth]{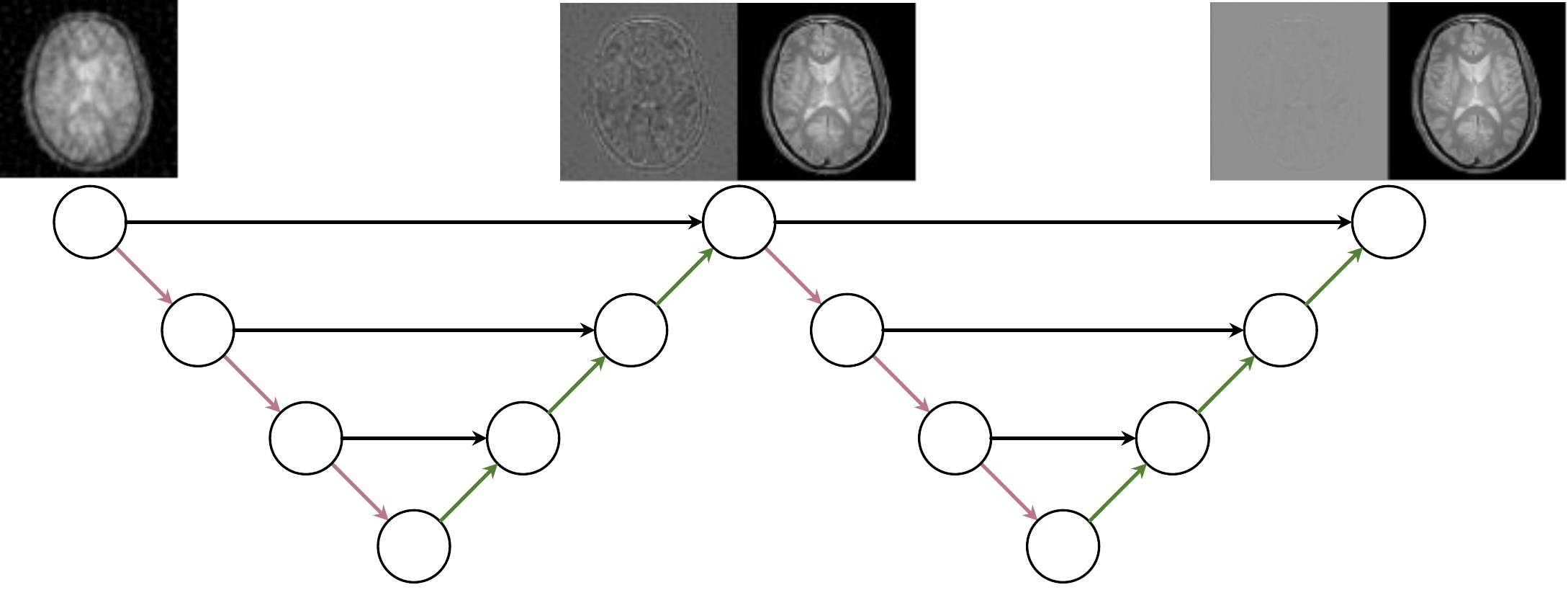}
	}
	\caption{One-fold (a) and two-fold architectures (b) of the generator $G$.}
	\label{fig:ReconRefineGAN}	
\end{figure*}

\subsection{Model Architecture}

In this section, we introduce the details of our neural network architecture, which is a variant of a convolutional autoencoder and deep residual network. 

\subsubsection{Fundamental blocks}

To begin discussing the model architecture, we first introduce three fundamental components in our generative adversarial model: \textit{encoder}, \textit{decoder}, with the insertion of the \textit{residual} block. 
The details of each block are described pictorially in Figure~\ref{fig:G}. 
The \textit{encoder} block, shaded in red, accepts a 4D tensor input and performs 2D convolution with \texttt{filter\_size} 3$\times$3, and \texttt{stride} is equal to 2 so that it performs down-sampling with convolution without a separate max-pooling layer.  
The number of feature maps \texttt{filter\_number} is denoted in the top. 
The \textit{decoder} block, shaded in green, functions as the convolution transpose, which enlarges the resolution of the 4D input tensor by two times. 
%
%
The \textit{residual} block, shaded in violet, is used to increase the depth of generator $G$ and discriminator $D$ networks, and it consists of three convolution layers: the first layer \texttt{conv\_i} with \texttt{filter\_size} is 3$\times$3, \texttt{stride} is 1 and the number of feature maps is \texttt{nb\_filters}, which reduces the dimension of the tensor by half. 
The second layer \texttt{conv\_m} performs the filtering via a 3$\times$3 convolution with the same number of feature maps \texttt{nb\_filters/2}. 
The remaining \texttt{conv\_o}, which has \texttt{filter\_size} 3$\times$3, \texttt{stride} is 1 and \texttt{nb\_filters}, feature maps, retrieves the input tensor shape so that they can be combined to form a residual bottleneck. 
This \textit{residual} block allows us to effectively construct a deeper generator $G$ and discriminator $D$ without suffering from the gradient vanishing problem~\cite{he_residual_2015}.


\subsubsection{Generator architecture}
Figure~\ref{fig:G} illustrates the architecture of our generator $G$. 
It is built based on the design of a convolutional autoencoder, which consists of an encoding path (left half of the network) to retrieve the compressed information in latent space, and a symmetric decoding path (right half of the network) that enables the prediction of synthesis. 
The convolution mode we used is ``same'', which leads the final reconstruction to have a size identical to the input images. 
The encoding and decoding paths consist of multiple levels, i.e., image resolutions, to extract features in different scales. 
Three types of introduced building blocks (i.e., \textit{encoder}, \textit{residual}, \textit{decoder} and their related parameters) are used to construct the proposed generator $G$.

It is worth noting that the proposed generator $G$ does not attempt to reconstruct the image directly. 
Instead, it is trained to produce the inverse amplitude of the noise caused by reconstruction from under-sampled data. 
The final reconstruction is obtained by adding the zero-filling input to the output from the generator $G$, which is similar to other current machine learning-based CS-MRI methods~\cite{lee2017deep, yu2017deep, mardani2017deep}.

\subsubsection{Discriminator architecture}
For the discriminator $D$, we use an architecture identical to that of the encoding path of the generator $G$. 
The output of the last \textit{residual} block is used to evaluate the mentioned adversarial loss $L_{adv}$ (Equation~\ref{eq:advloss}). 
To reiterate, if the discriminator receives the image $s \in S$, it will result in $D\left( s \right) = 1$, as a true result.
Otherwise, the reconstruction $\overline s$ will be recognized as a fake result, or equivalently, $D\left( {\overline s } \right) = D\left( {G\left( s_0 \right)} \right) = 0$.


\subsection{Full Objective Function and Training Specifications}
In summary, our system involves two sub-networks which are trained adversarially to minimize the following loss:
\begin{equation}
\begin{aligned}
{L_{total}} = {L_{adv}}\left( {G,D} \right) + \alpha {L_{freq}}\left( G \right) + \gamma {L_{imag}}\left( G \right)
\end{aligned}
\end{equation}
where $\alpha$ and $\gamma$ are the weights which help to control the balance between each contribution. 
We set $\alpha=1.0$ and $\gamma=10.0$ for all the experiments. 
The Adam optimizer~\cite{kingma2014adam} is used with the initial learning rate of $1e^{-4}$, which decreases monotonically over 500 epochs. 
Our source code will be tentatively published and available~\footnote{\url{http://hvcl.unist.ac.kr/RefineGAN/}}. 
The entire framework was implemented using a system-oriented programming wrapper (tensorpack~\footnote{\url{http://tensorpack.readthedocs.io/}}) of the tensorflow~\footnote{\url{http://www.tensorflow.org/}} library.

\subsection{Chaining with Refinement Network}
The proposed generator $G$ by itself can perform an end-to-end reconstruction from the zero-filling MRI to the final prediction. 
However, in the real-world setup, many iterative methods also take extra steps to go through the current result and then attempt to ``correct'' the mistakes. 
Therefore, we introduce an additional step in refining the reconstruction by concatenating a chain of multiple generators to resolve the ambiguities of the initial prediction from the generator. 
For example, Figure~\ref{fig:ReconRefineGAN} shows the two-fold chaining generator in a self-correcting form. 
By forcing the desired ground-truth between them, the entire solution becomes a target-driven approach. 
This enables our method to be a single-input and multi-output model, where each checkpoint in between attempts to produce better a reconstruction. 
Because the architecture of each sub-generator is the same, we can think of the proposed model is another variant of the recurrent neural network that treats the entire sub-generator as a single state without sharing the weights after unfolding. 
The loss training curves of the checkpoints decrease as the number of checkpoints increases, and they eventually converge as the number of training iterations (epochs) increases. 
Interestingly, our generator can be considered a V-cycle in the multigrid method that is commonly used in numerical analysis, where the contraction in the encoding path is similar to \textit{restriction} heads from fine to coarse grid. 
The expansion in the decoding path spans along the \textit{prolongation} toward the final reconstruction, and the skip connections act as the \textit{relaxation}. 
We refer to the first check point as \textit{ReconGAN} and the second check point in our two-fold chaining network as \textit{RefineGAN}. 
\rv{
The entire chaining structure (with 2 generators) is trained together. 
Each individual cycle has its own variable scope and hence, their weight are updated differently. 
The later structure serves as a boosting module which improves results.
}

%% file: result.tex
\label{sec:result}

\subsection{Results on real-valued MRI data}

\begin{figure}[ht]
	\centering
	\includegraphics[width=0.9\linewidth]{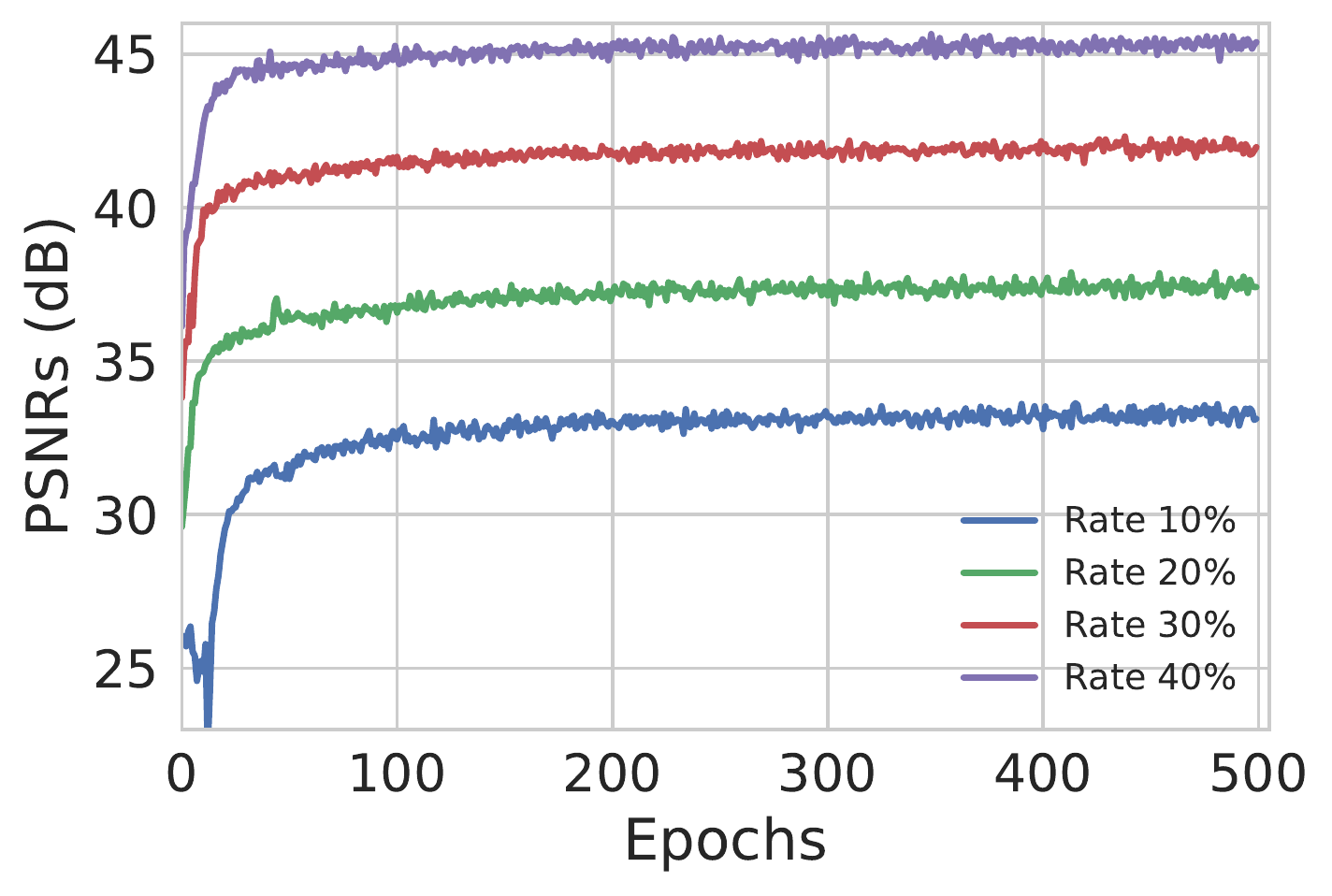}
	\caption{PSNR curves of RefineGAN with different undersampling rates on the brain training set over 500 epochs.}
	\label{fig:training_curve}	
\end{figure}

We trained many versions of the proposed networks with different factors of under-sampling masks (10\%, 20\%, 30\%, 40\%) on the training sets. 
As shown in Figure~\ref{fig:training_curve}, those trained models (for RefineGAN) reach a convergence stage in a few hundred training iterations. 
The performances on the test sets are expected to show similar results. 

\begin{figure}[ht]
	\centering
	\subfloat[10\%]
	{
		\includegraphics[width=0.22\linewidth]{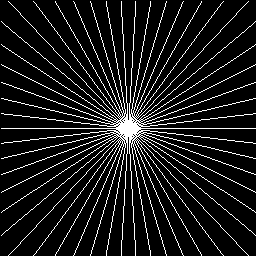}
	}
	\subfloat[20\%]
	{
		\includegraphics[width=0.22\linewidth]{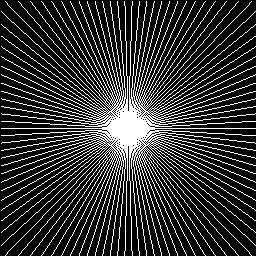}
	}
	\subfloat[30\%]
	{
		\includegraphics[width=0.22\linewidth]{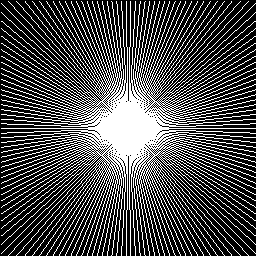}
	}
	\subfloat[40\%]
	{
		\includegraphics[width=0.22\linewidth]{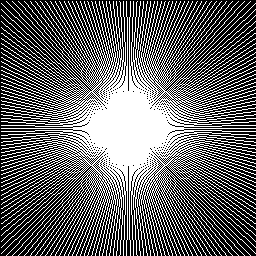}
	}
	\caption{Radial sampling masks used in our experiments.}
	\label{fig:masks}	
\end{figure}

\begin{table}[!ht]
	\sisetup{round-mode=places,round-precision=4}
	\centering
	\caption{Running time comparison of various CS-MRI methods on two testing datasets (in seconds).}
	\begin{tabular}{|c|c|r|r|}
		\hline
		Abbv. & \multicolumn{1}{c|}{Methods} & \multicolumn{1}{c|}{Brain} & \multicolumn{1}{c|}{Chest} \\
		\hline
		CSCMRI & \multicolumn{1}{c|}{~\cite{quanisbi, quan2016miccai}} & 8.56808 & 9.37082 \\
		\hline
		DLMRI & \multicolumn{1}{c|}{~\cite{ravishankar2011mr, caballero2012dictionary, caballero2014dictionary}} & 604.24623 & 613.84531 \\
		\hline
		DeepADMM & \multicolumn{1}{c|}{ ~\cite{sun2016deep}} & 0.31725 & 0.28677 \\
		\hline
		\rv{DeepCascade} & \multicolumn{1}{c|}{ ~\cite{schlempler2017deepcascade}}  & 0.22182 & 0.25627\\
		\hline
		SingleGAN & \multicolumn{1}{c|}{ ~\cite{yu2017deep, mardani2017deep}} & \rv{0.064599} & \rv{0.075529}\\
		\hline
		ReconGAN &   ---     & \rv{0.060753} & \rv{0.068871} \\
		\hline
		RefineGAN &  ---     & \rv{0.106157} & \rv{0.111607}\\
		\hline
	\end{tabular}%
	\label{tab:methods}%
\end{table}%

\begin{figure}[]
	\vspace{-0.3in}	
	\centering
	\subfloat[Brain 10\%]
	{
		\includegraphics[width=0.24\textwidth]{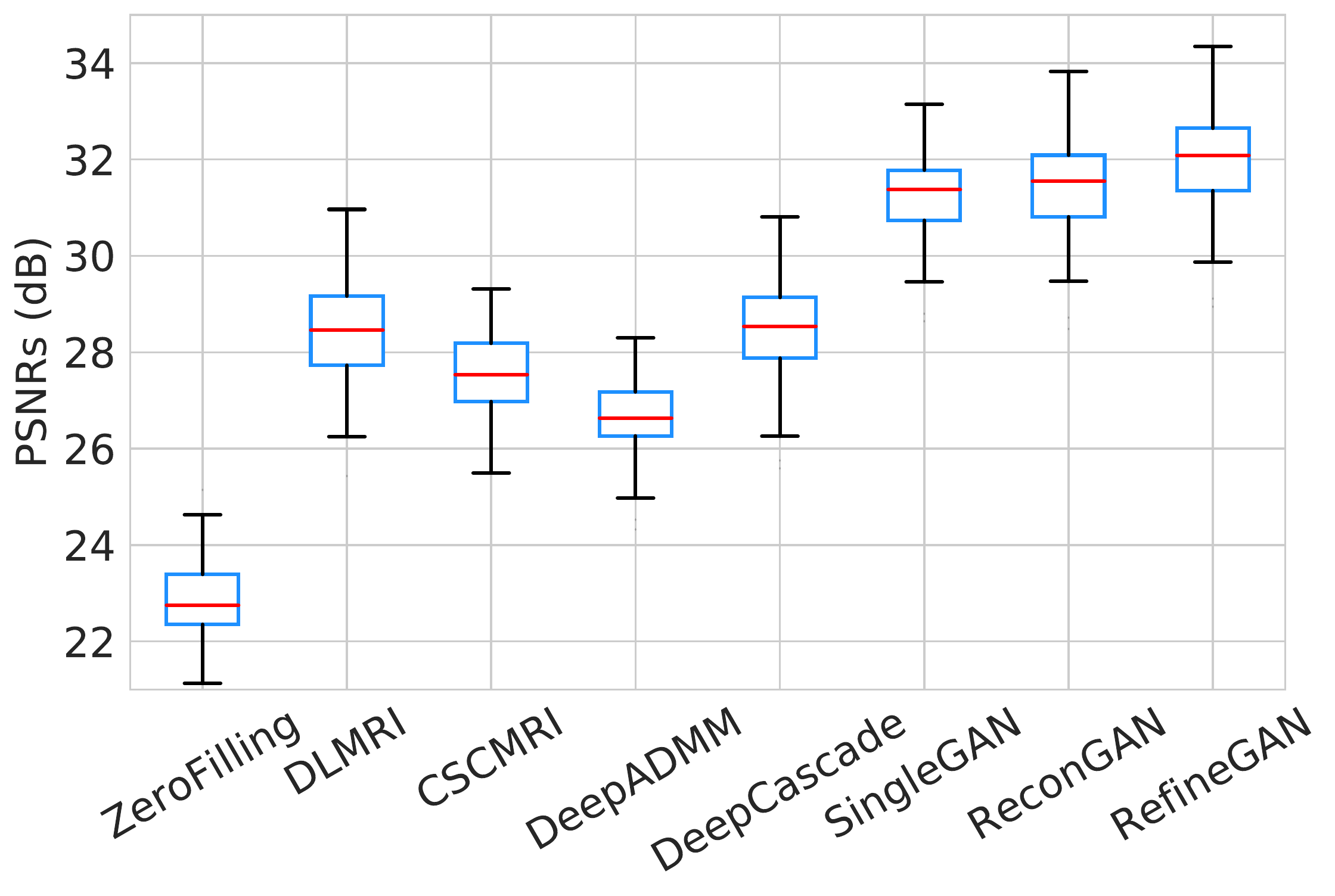}
	}	
	\subfloat[Chest 10\%]
	{
		\includegraphics[width=0.24\textwidth]{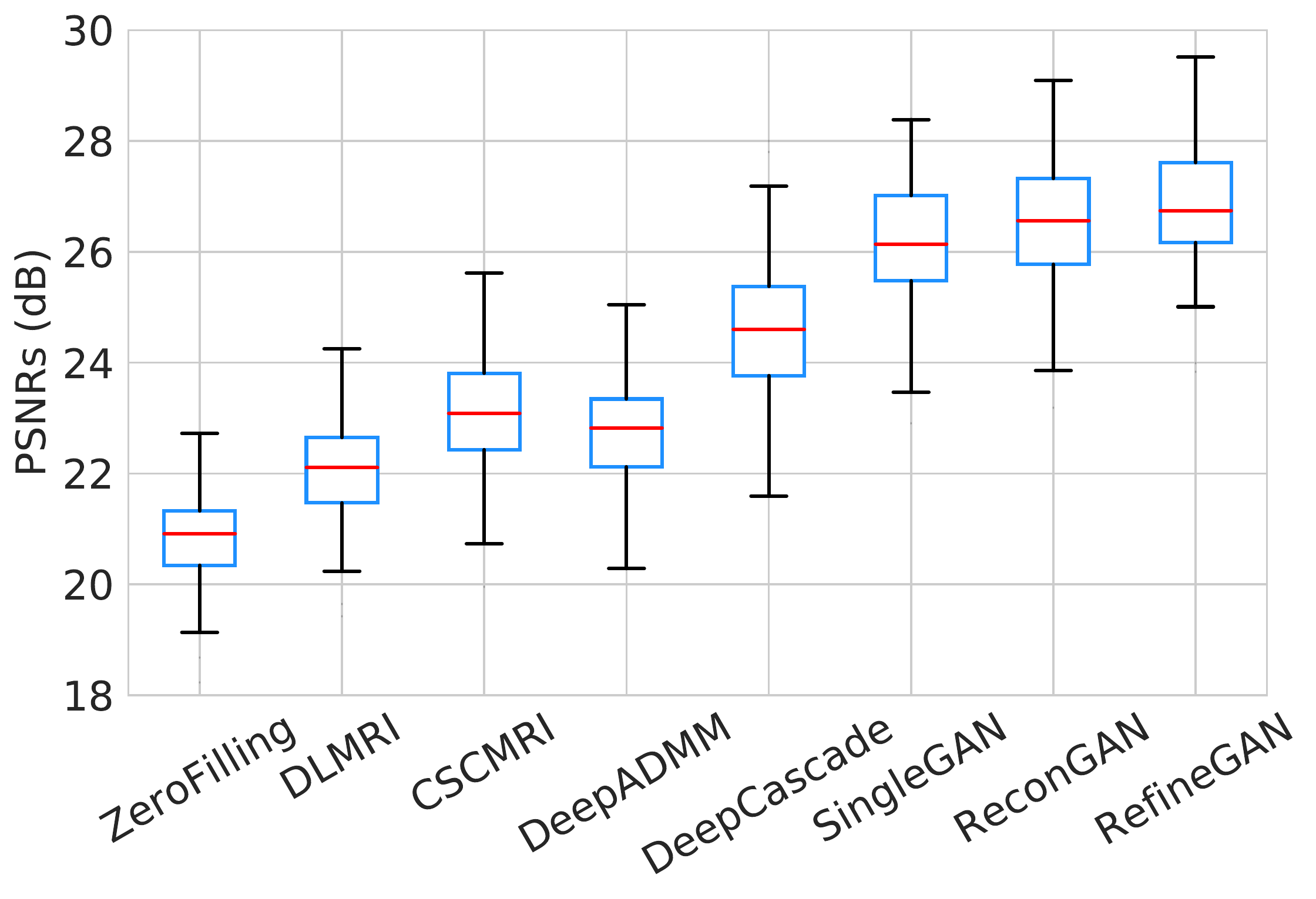}
	}	\\
	\vspace{-0.10in}
	\subfloat[Brain 30\%]
	{
		\includegraphics[width=0.24\textwidth]{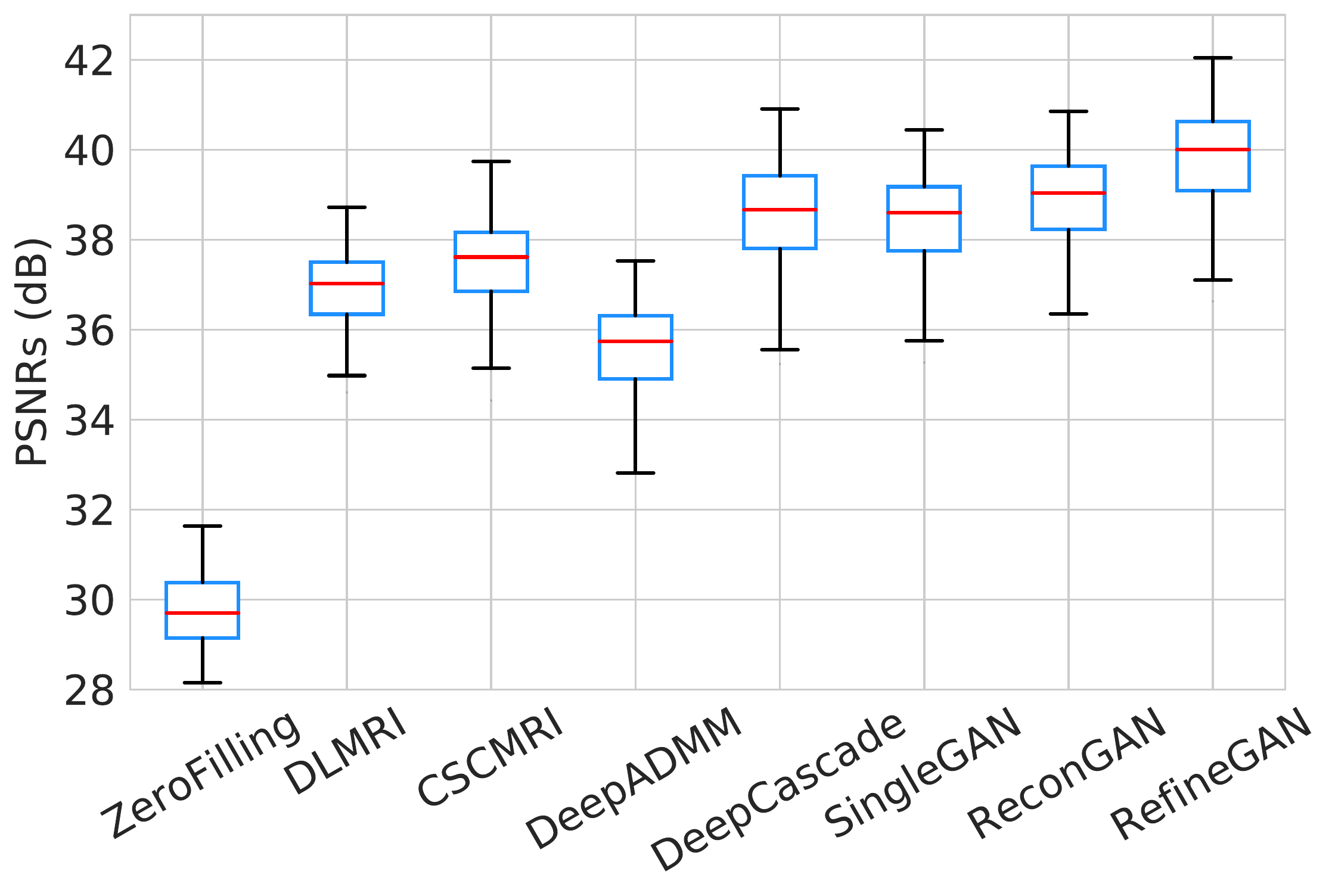}
	}	
	\subfloat[Chest 30\%]
	{
		\includegraphics[width=0.24\textwidth]{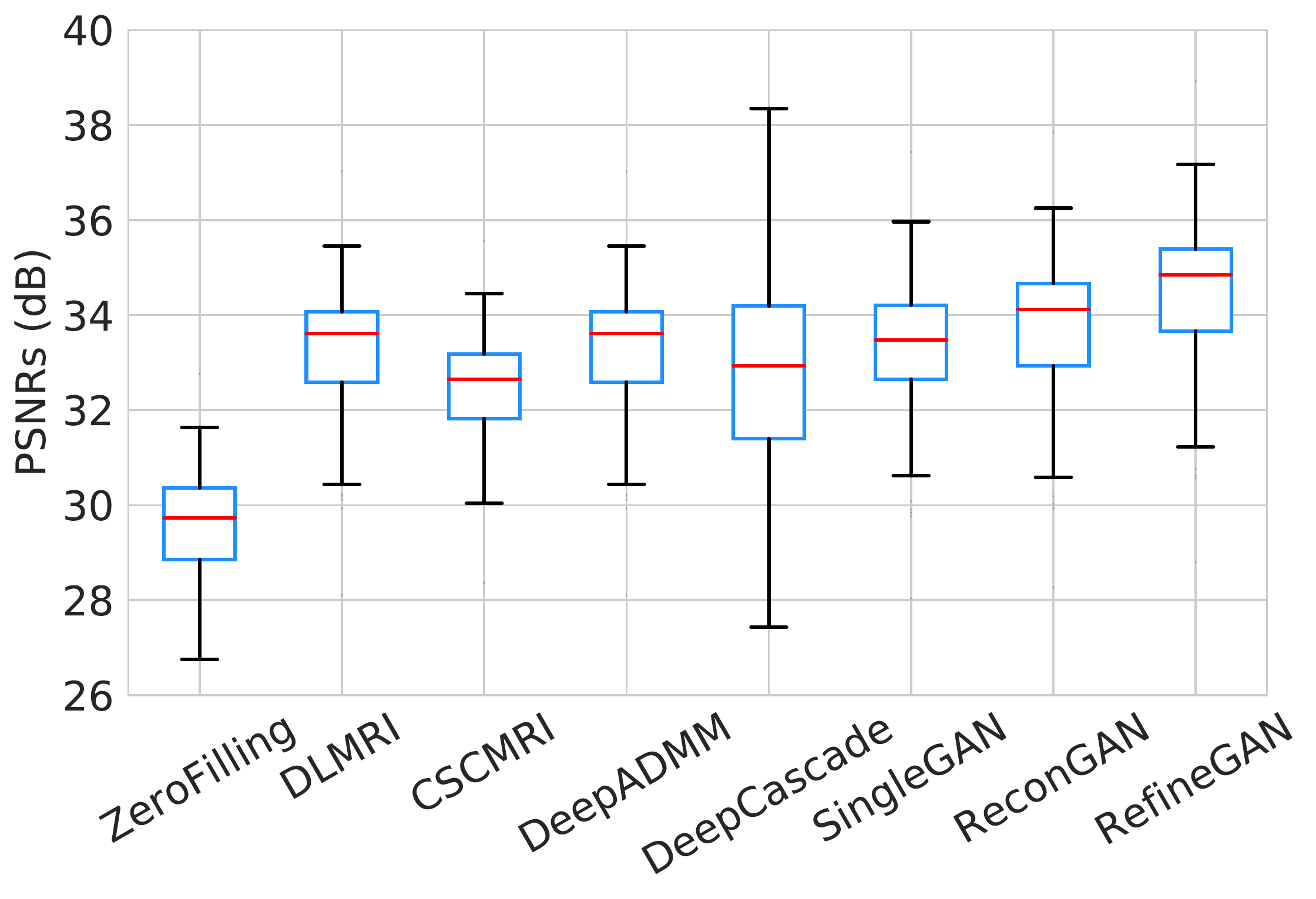}
	}
	\caption{PSNRs evaluation on the brain and chest test set. Unit: dB}
	\vspace{-0.20in}
	\label{fig:psnr}	
\end{figure}
\begin{figure}[]
	\centering
	\vspace{-0.25in}
	\subfloat[Brain 10\%]
	{
		\includegraphics[width=0.24\textwidth]{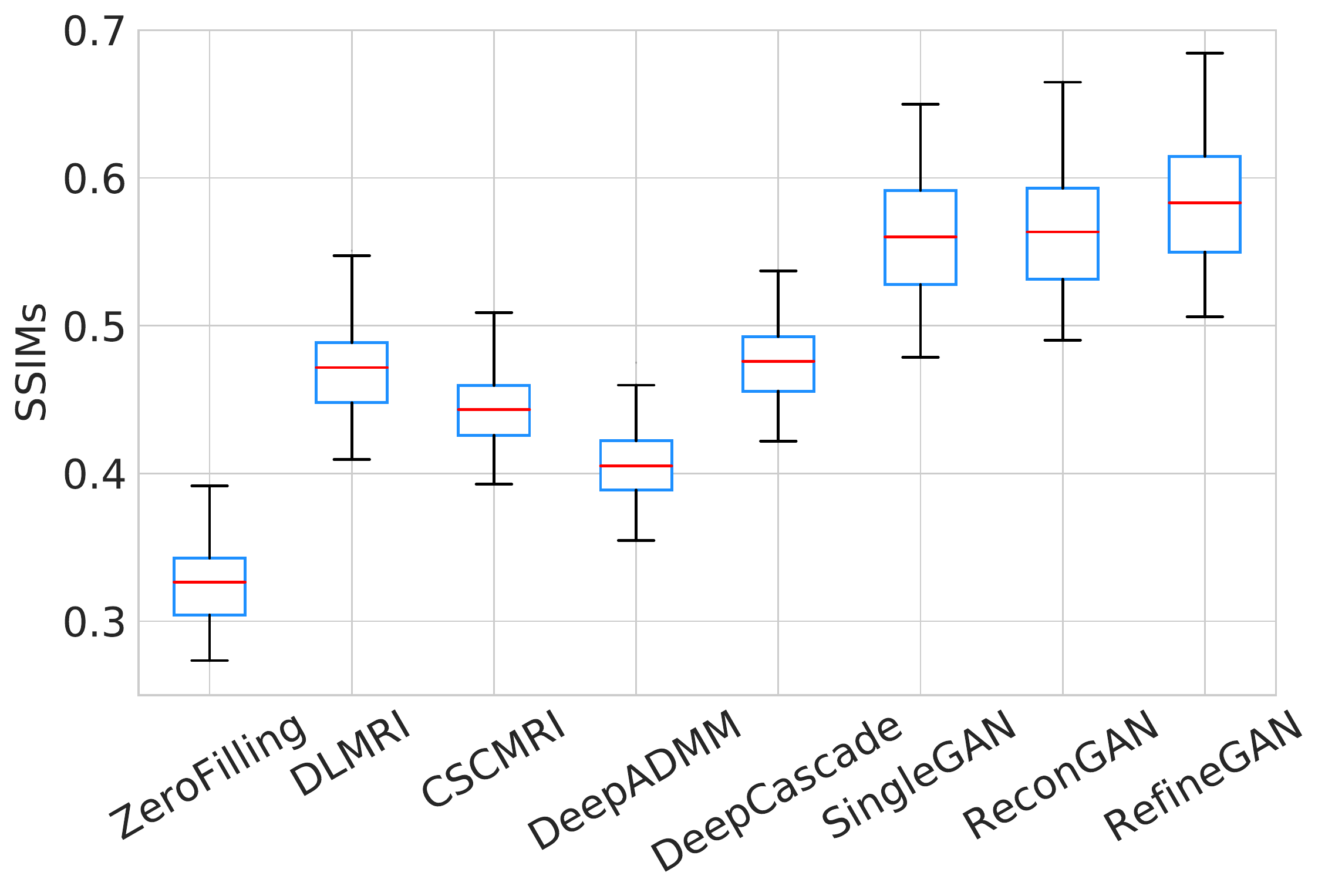}
	}	
	\subfloat[Chest 10\%]
	{
		\includegraphics[width=0.24\textwidth]{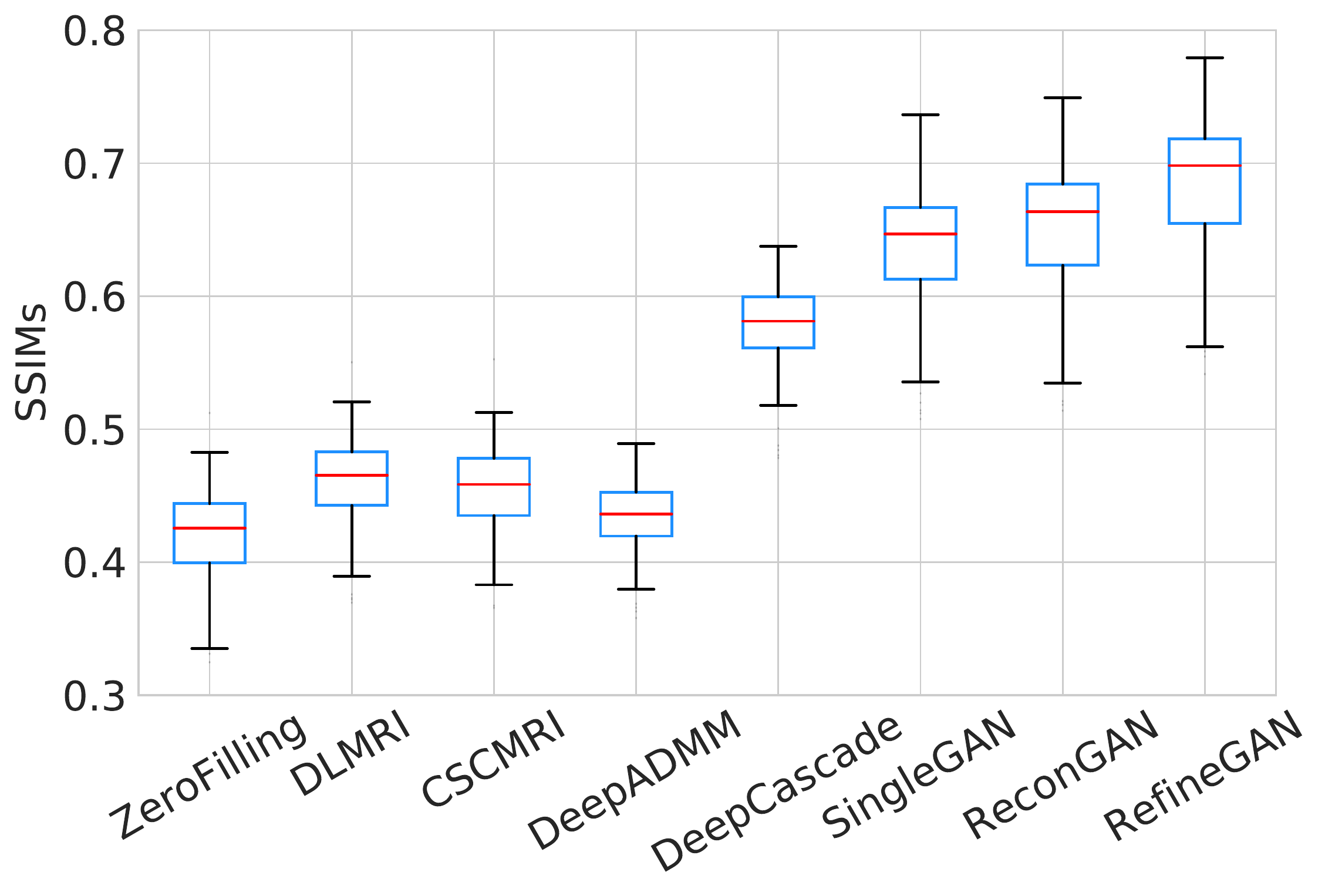}
	}\\	
	\vspace{-0.10in}
	\subfloat[Brain 30\%]
	{
		\includegraphics[width=0.24\textwidth]{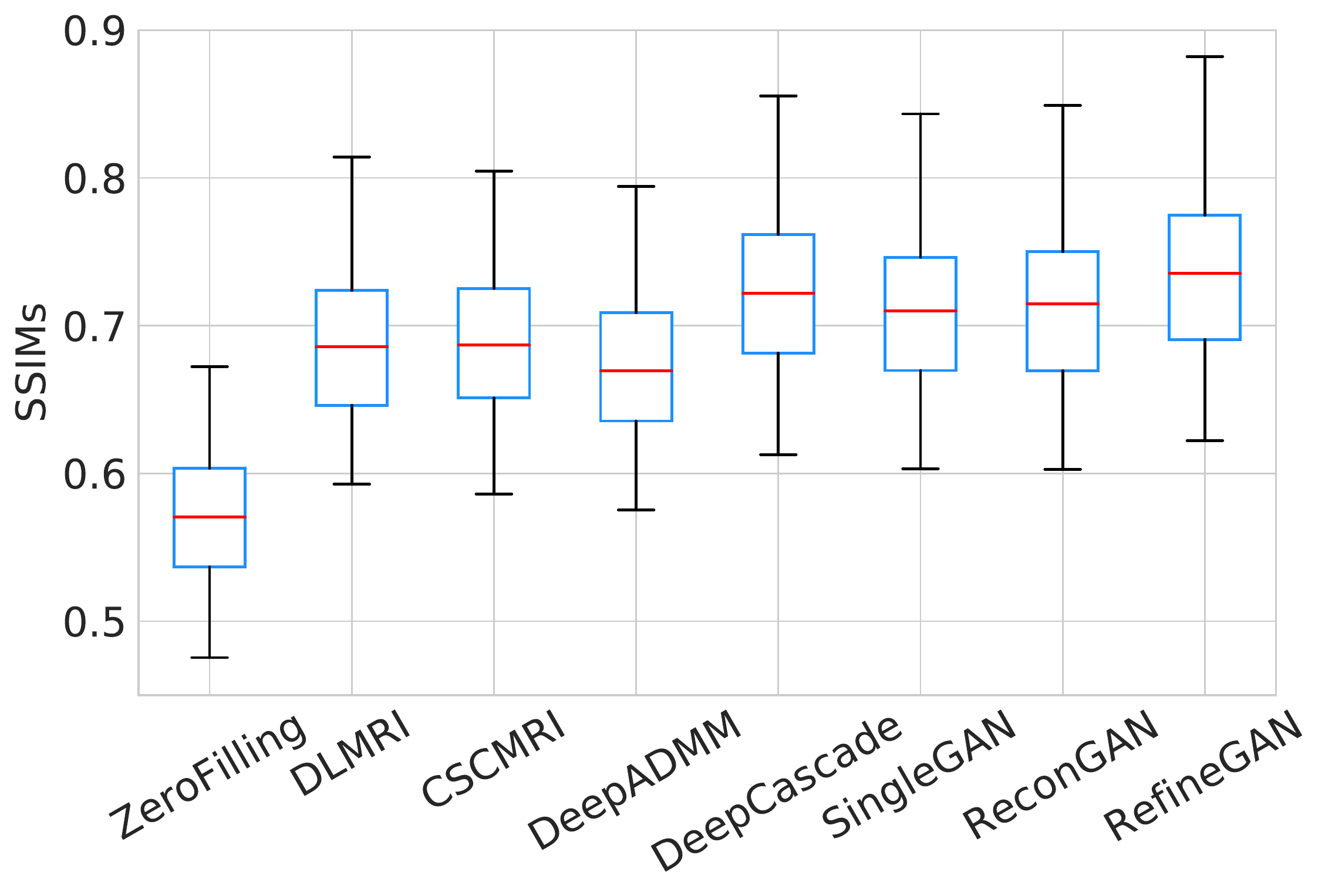}
	}	
	\subfloat[Chest 30\%]
	{
		\includegraphics[width=0.24\textwidth]{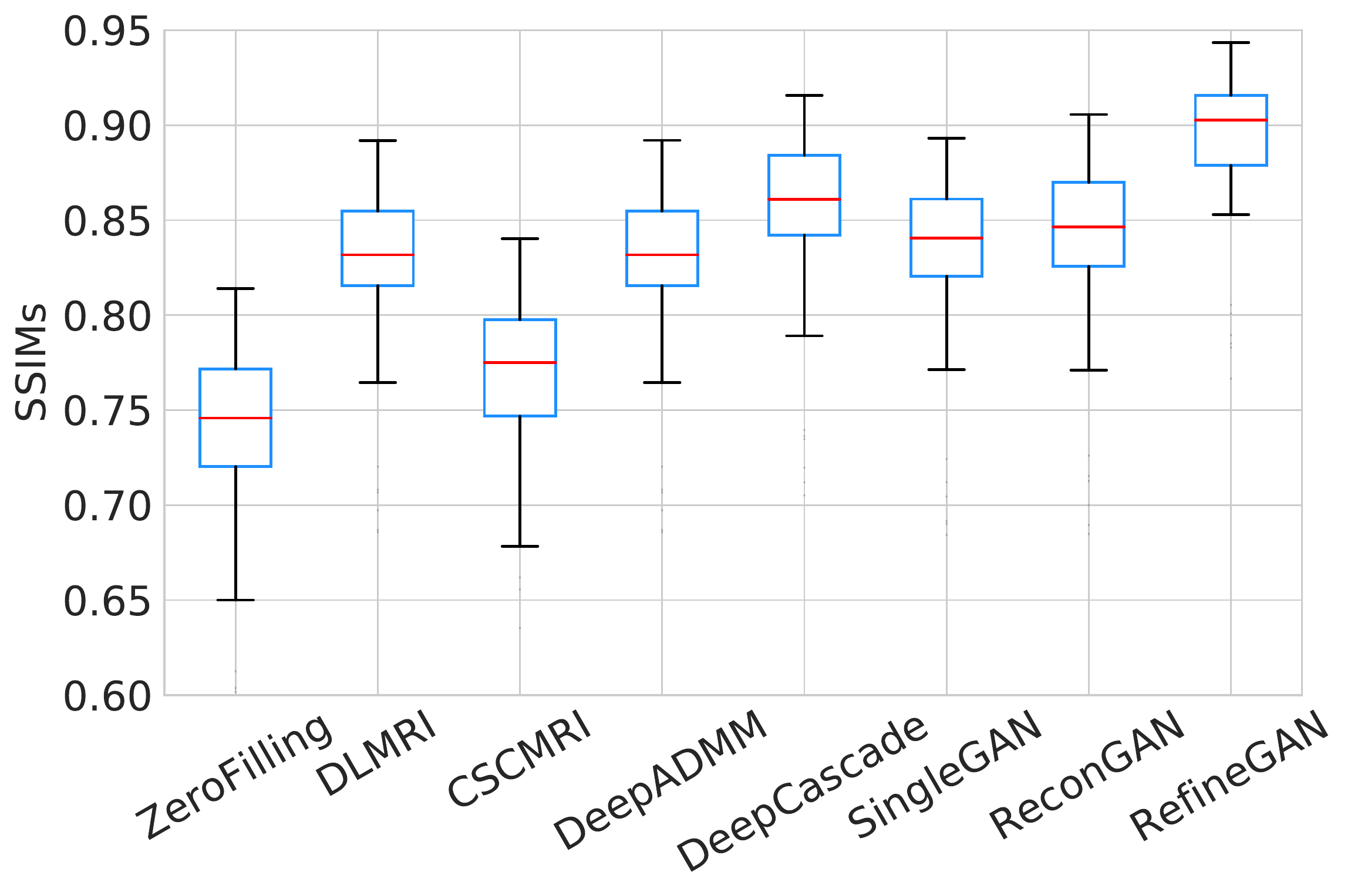}
	}
	\caption{SSIMs evaluation on the brain and chest test set}
	\label{fig:ssim}
	\vspace{-0.20in}	
\end{figure}

\begin{figure}[]
	\centering
	\vspace{-0.3in}	
	\subfloat[Brain 10\%]
	{
		\includegraphics[width=0.24\textwidth]{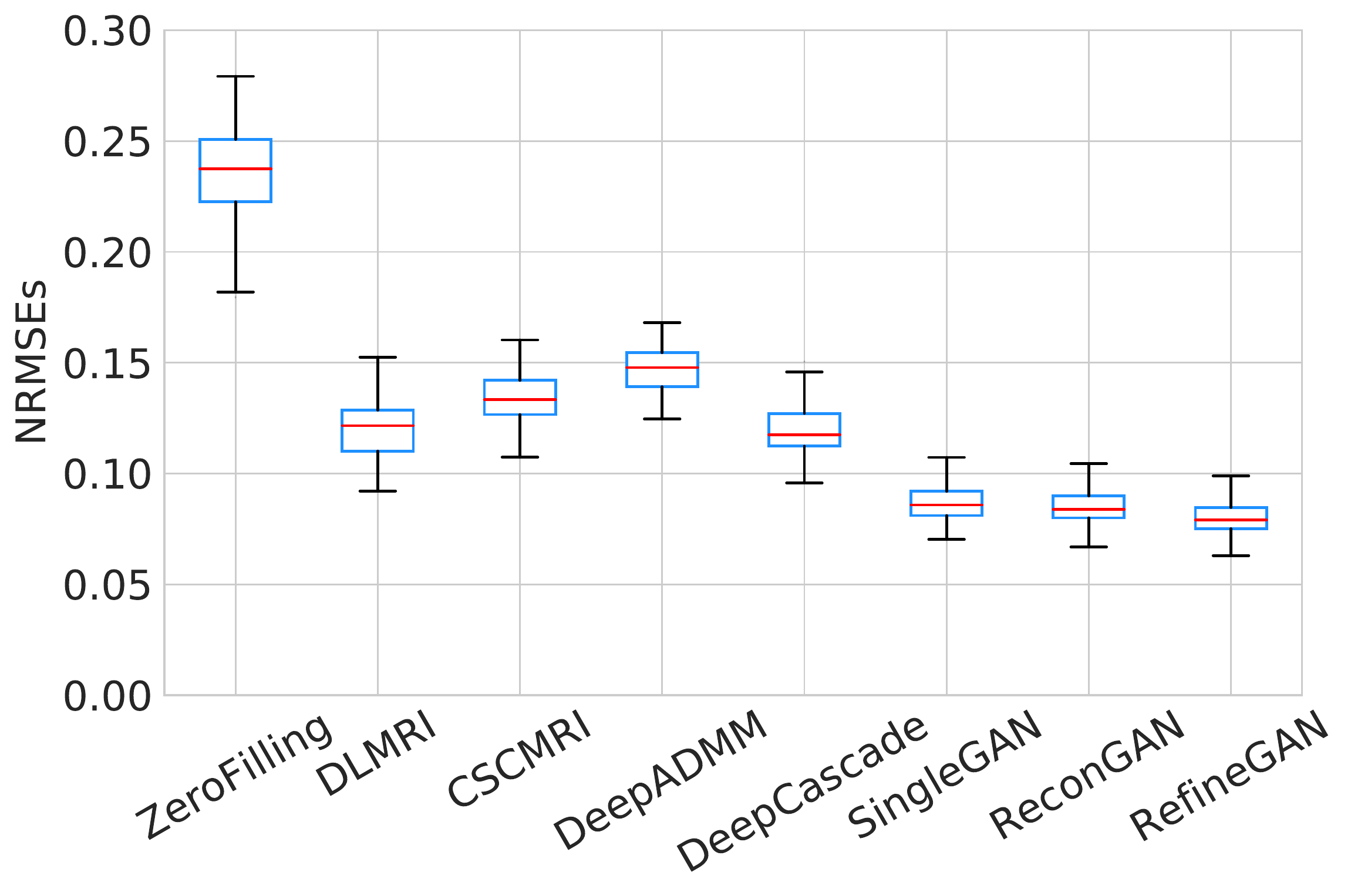}
	}	
	\subfloat[Chest 10\%]
	{
		\includegraphics[width=0.24\textwidth]{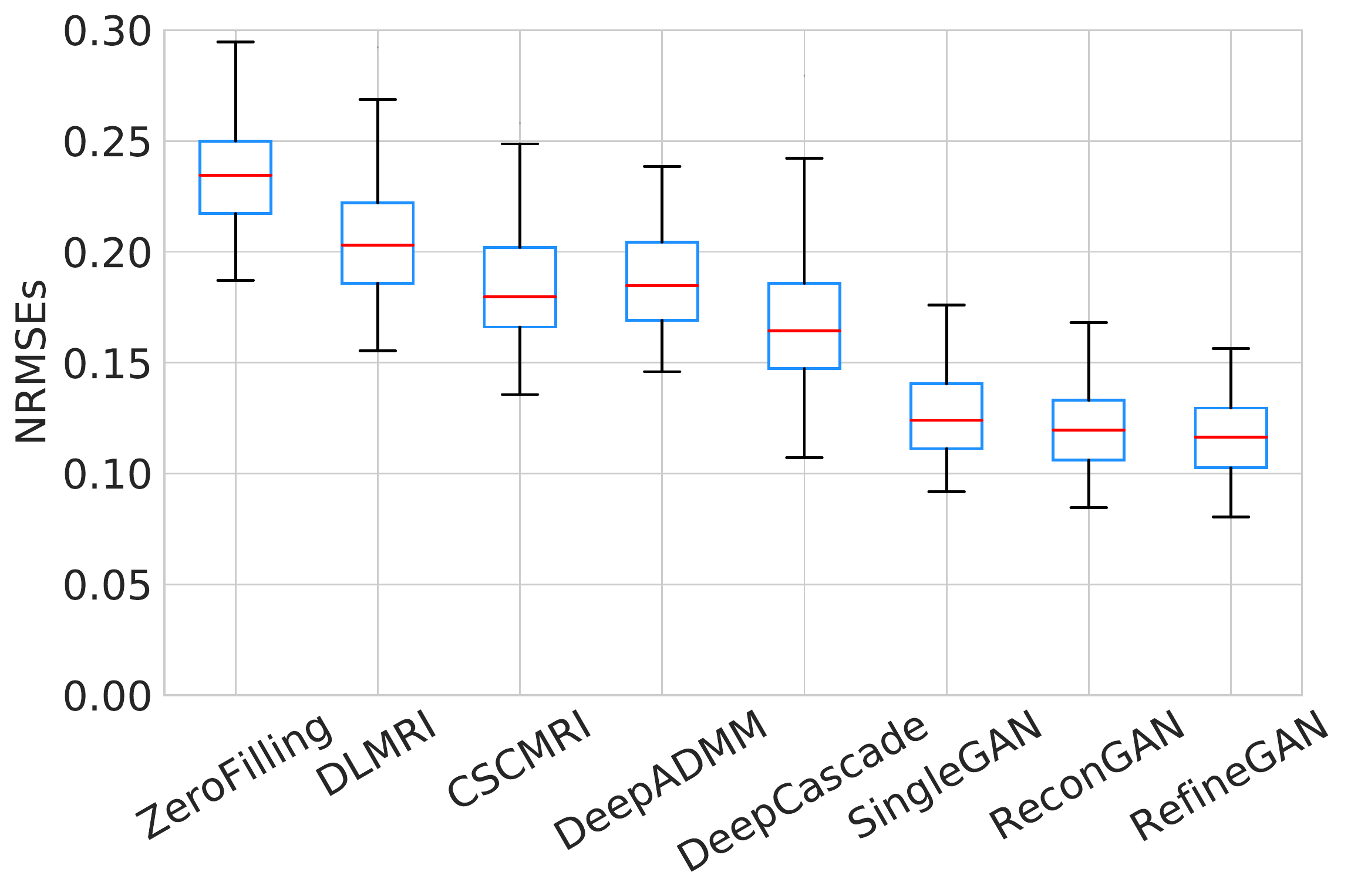}
	}	\\
	\vspace{-0.10in}
	\subfloat[Brain 30\%]
	{
		\includegraphics[width=0.24\textwidth]{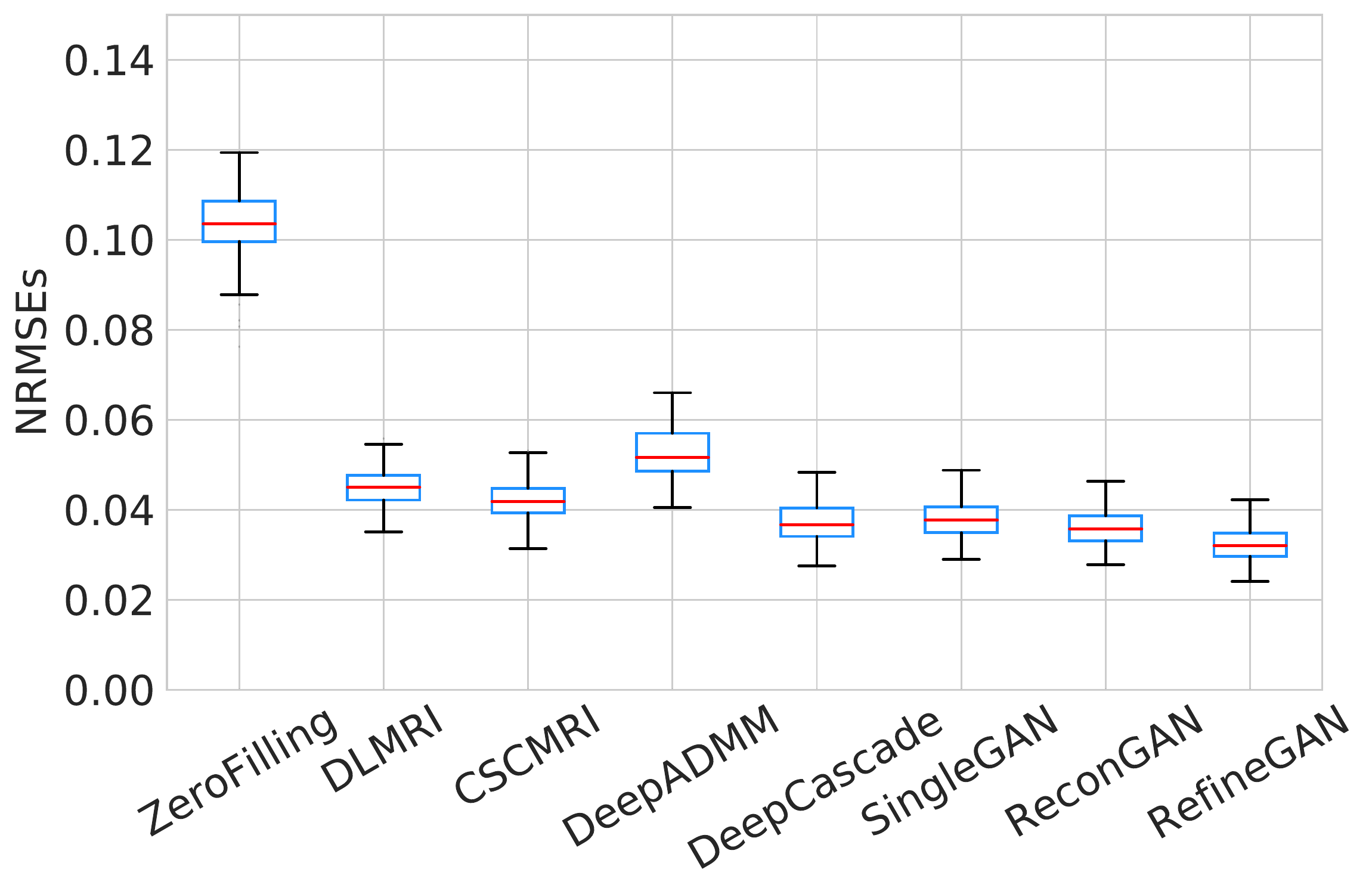}
	}	
	\subfloat[Chest 30\%]
	{
		\includegraphics[width=0.24\textwidth]{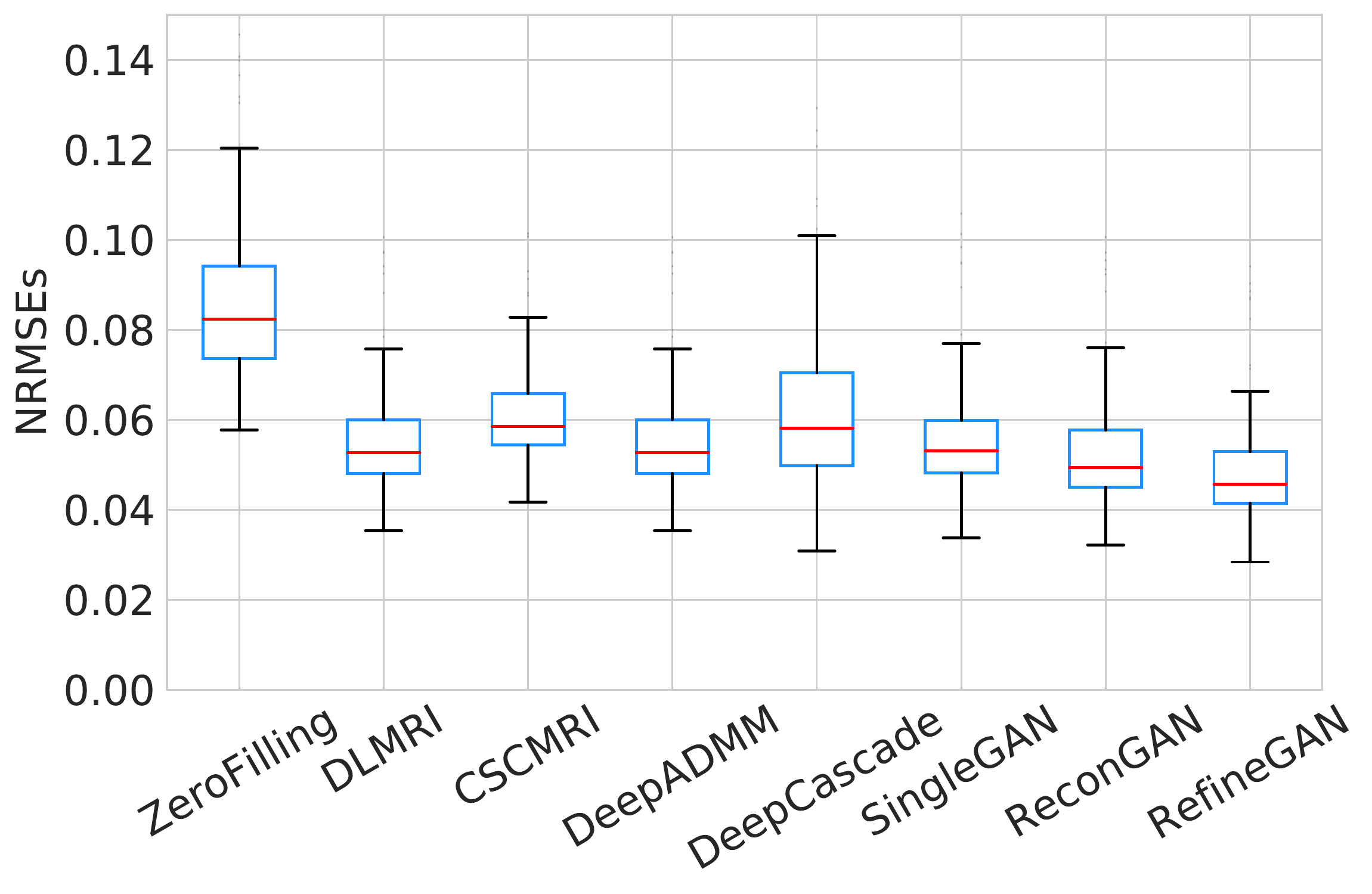}
	}
	\caption{NRMSEs evaluation on the brain and chest test set}
	\label{fig:nrmse}	
	\vspace{-0.25in}
\end{figure}

\begin{figure*}[]
	\centering
	\vspace{-0.2in}
	\subfloat[]
	{\includegraphics[width=0.95\linewidth]{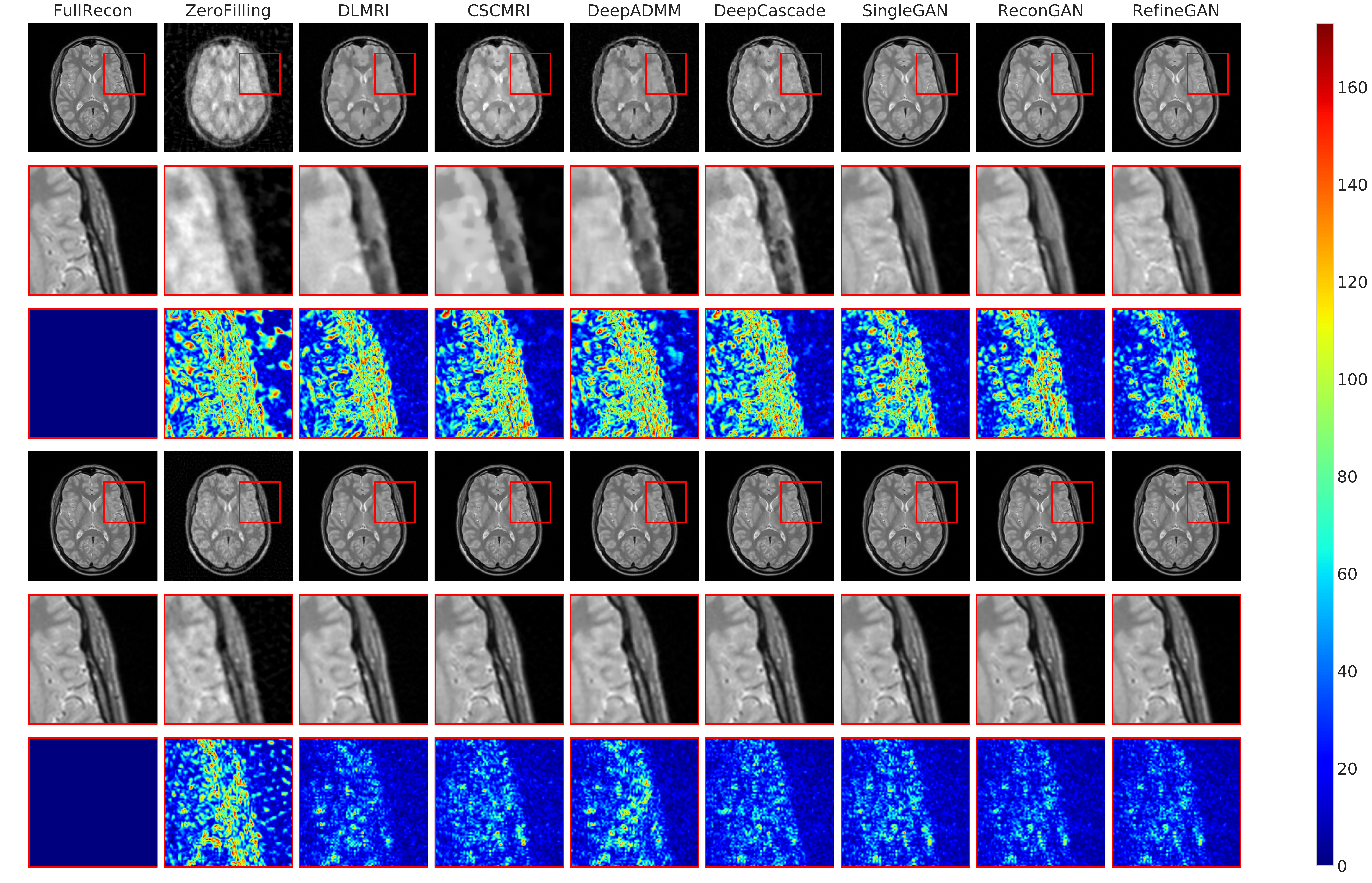}}
	\\
	\vspace{-0.1in}
	\subfloat[]
	{\includegraphics[width=0.95\linewidth]{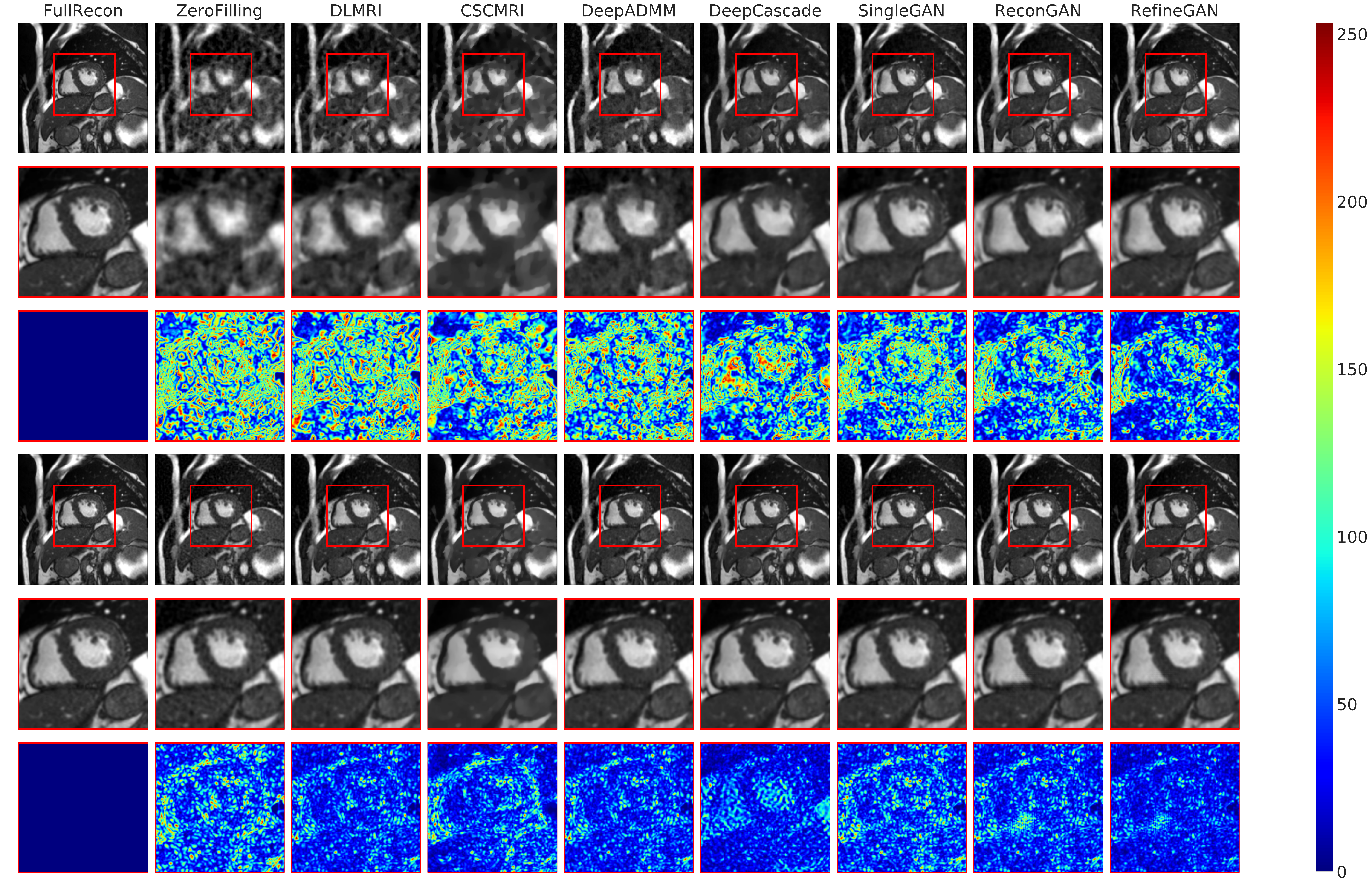}}
	\caption{Image quality comparison on the brain (a) and chest dataset (b) at a sampling rate 10\% (top 3 rows) and 30\% (bottom 3 rows): Reconstruction image, zoom in result and 10$\times$ error map compared to the full reconstruction.}
	\label{fig:brain_chest_zoom}	
\end{figure*}

%

%
We used two sets of MR images 
from the IXI database\footnote{\url{http://brain-development.org/ixi-dataset/}} (the brain dataset)
and from the Data Science Bowl challenge \footnote{\url{https://www.kaggle.com/c/second-annual-data-science-bowl/data}} (the chest dataset) to assess the performance of our method by comparing our results with state-of-the-art CS-MRI methods (e.g., Convolutional sparse coding-based~\cite{quanisbi, quan2016miccai}, patch-based dictionary~\cite{ravishankar2011mr, caballero2012dictionary, caballero2014dictionary}, deep learning-based~\cite{sun2016deep, wang2016accelerating, lee2017deep, schlempler2017deepcascade}, and GAN-based~\cite{yu2017deep, mardani2017deep}).
%
The image resolution of each image is 256x256.
From each database, we randomly selected 100 images for training the network and another 100 images for testing (validating) the result. 
We conducted the experiments for various sampling rates (i.e., 10\%, 20\%, 30\%, and 40\% of the original $k$-space data), corresponding to 10$\times$, 5$\times$, 3.3$\times$, and 2.5$\times$ factors of acceleration. 
We assume the target MRI data type is static, and radial sampling masks are applied (Figure~\ref{fig:masks}). 
It is worth noting that our experimental data are real-valued MRI images, which require pre-processing of the actual acquisition from the MRI scanner because the actual MRI data is complex-valued.
Additional data preparation steps, such as data range normalization and imaginary channel concatenation, are also required.

\textbf{Running Time Evaluation:} Table~\ref{tab:methods} summarizes the running times of our method and other state-of-the-art learning-based CS-MRI methods. 
Even though dictionary learning-based approaches leverage pre-trained dictionaries, their reconstruction time depends on the numerical methods used. 
For example, CSCMRI by Quan and Jeong~\cite{quanisbi, quan2016miccai} employed a GPU-based ADMM method, which is considered one of the state-of-the-art numerical methods, but the running time is still far from interactive (about 9 seconds). 
Another type of dictionary learning-based method, DLMRI~\cite{ravishankar2011mr, caballero2012dictionary, caballero2014dictionary}, solely relies on the CPU implementation of a greedy algorithm, so their reconstruction times are significantly longer (around 600 seconds) than those of the others with GPU-acceleration. 
Deep learning-based methods, including DeepADMM, DeepCascade, and our method, are extremely fast (e.g., less than a second) because deploying a feed-forward convolutional neural network is a single-pass image processing that can be accelerated using GPUs reasonably well. 
DeepADMM significantly accelerated time-consuming iterative computation to as low as 0.2 second. 
The running times of SingleGAN~\cite{yu2017deep, mardani2017deep} and our ReconGAN are, all similarly, about 0.07 second because they share the same network architecture (i.e., single-fold generator $G$). 
%
The running time of RefineGAN 
is about twice as long because two identical generators are serially chained in a single architecture, \rv{but it still runs at an interactive rate (around 0.1 second).} 
As shown in this experiment, we observed that deep learning-based approaches are well-suited for CS-MRI in a time-critical clinical setup due to their extremely low running times.

%
%
%
%

%
\textbf{Image Quality Evaluation:} 
To assess the quality of reconstructed images, we use three image quality metrics, such as Peak-Signal-To-Noise ratio (PSNR), Structural Similarity (SSIM) \rv{and Normalized root-mean-square error (NRMSE)}
Figure~\ref{fig:psnr},~\ref{fig:ssim} and~\ref{fig:nrmse} show their PSNRs, SSIMs \rv{and NRMSEs} error graphs, respectively. 
Additionally, Figure~\ref{fig:brain_chest_zoom} shows the representative reconstruction of the brain and chest test sets, respectively, using various reconstruction methods at different sampling rates (10\% and 30\%) and their 10$\times$ magnified error plots using a jet color map (blue: low, red: high error). 
%
Overall, our methods (ReconGAN and RefineGAN) are able to reconstruct images with better PSNRs, SSIMs \rv{and NRMSEs}. 
Note that we used the identical generator and discriminator networks (i.e., the same number of neurons) for \del{DeepDirect,} SingleGAN, and our own method for a fair comparison.      
We observed that our cyclic loss increases the PSNR by around 1dB, and the refinement network further reduces the error to a similar degree.
\stkout{
We also observed that our cyclic loss seems more helpful for higher sampling rates (see the difference between SingleGAN and ReconGAN) which is due to data consistency constraints enforced by the loss. 
We think that for lower sampling rates, not much information can be interpolated from the data, so the final results become similar over different methods, mostly synthesized from learned information. 
However, for higher sampling rates, our method can accurately interpolate the input sampling data, which eventually improves the total image quality to close to that of the full reconstruction.
}

By qualitatively comparing the reconstructed results, we found that deep learning-based methods generate more natural images than dictionary-based methods. 
For example, CSCMRI and DLMRI produce cartoon-like piecewise linear images with sharp edges, which is mostly due to sparsity enforcement. 
In comparison, our method generates results that are much closer to full reconstructions while edges are still preserved; in addition, noise is significantly reduced. 
Note also that, comparing to the other CS-MRI methods, our method can generate superior results especially at extremely low sampling rates (as low as 10\%, see Figure~\ref{fig:brain_chest_zoom}).

\begin{figure*}[]
	\centering
	{\includegraphics[width=1\linewidth]{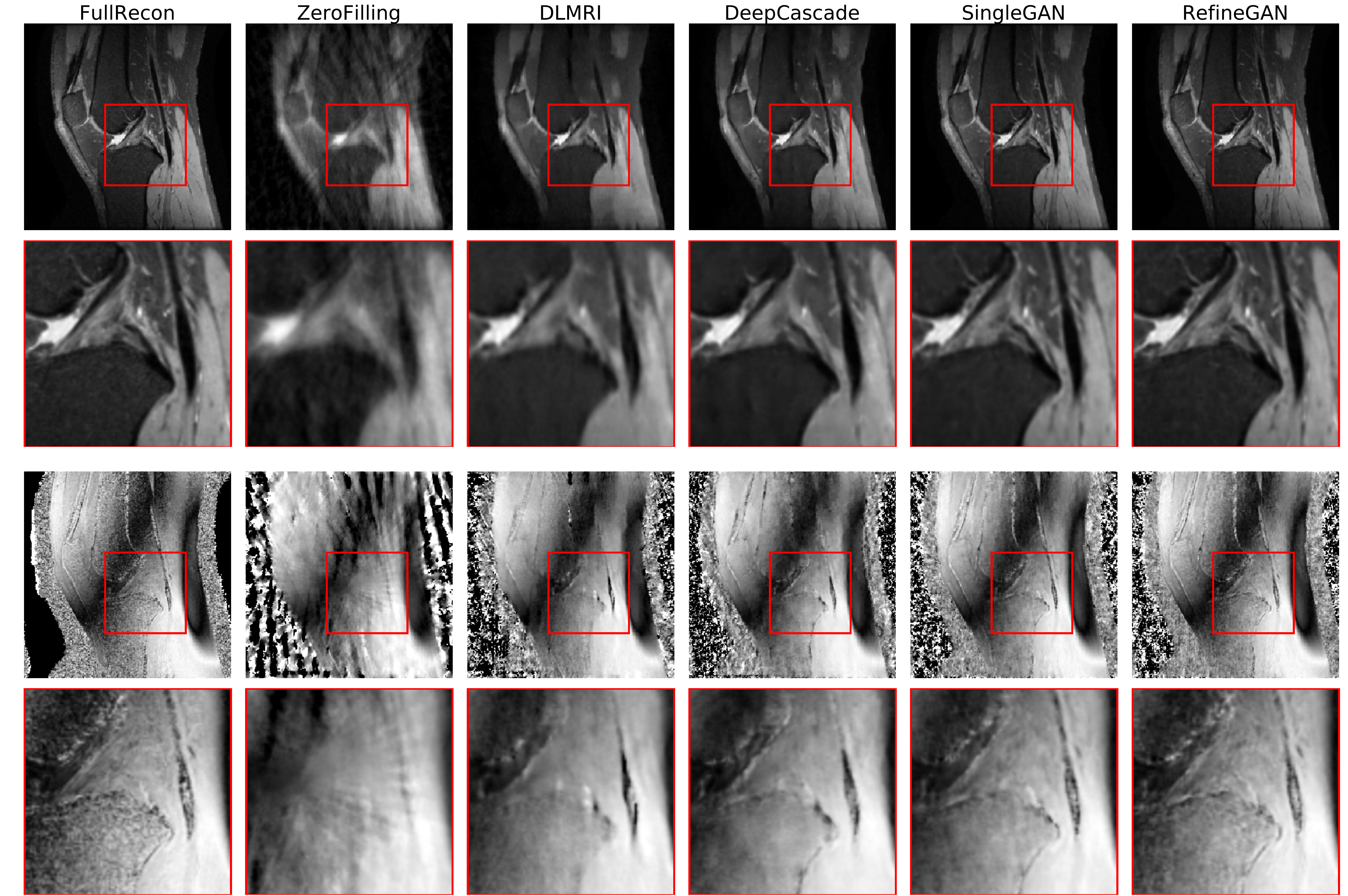}}
	\caption{Image quality comparison on the knees dataset (top 2 rows: magnitude images, and bottom 2 rows: phase images) at a sampling rate 10\% : Reconstruction images and zoom-in results}
	\label{fig:knee_zoom}	
\end{figure*}

\begin{figure}[tbp]
	\centering
	\includegraphics[width=1\linewidth]{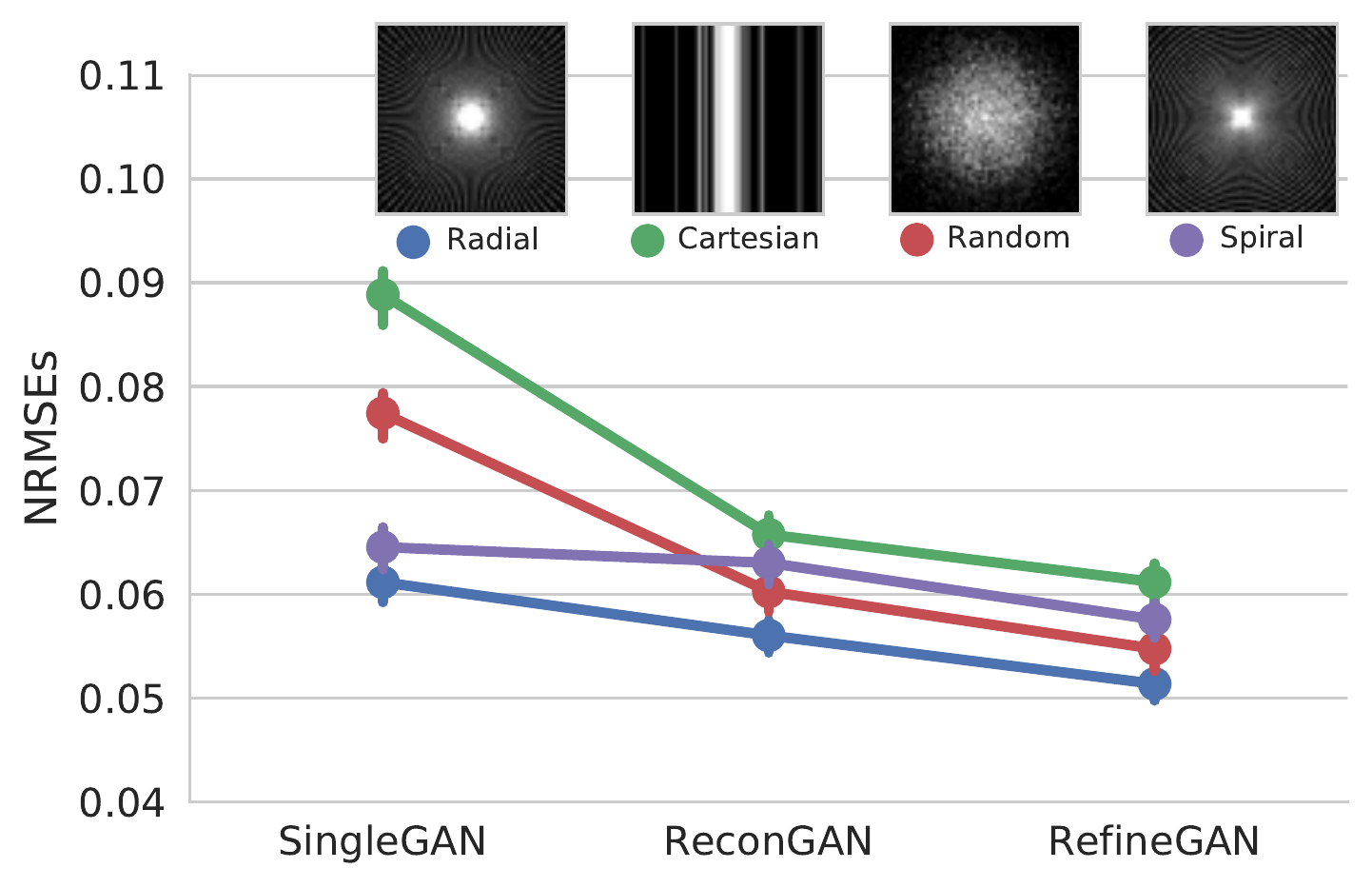}
	\caption{NRMSEs evaluation on the knees test set at sampling rate 20\% with various sampling masks}
	\label{fig:nrmse_various_2}	
\end{figure}

\rv{
\subsection{Results on complex-valued MRI data}
The proposed method can accept 2-channel complex-valued zero-filling image as an input and return a 2-channel complex-valued reconstruction without loss of generality. 
We used another public database of MR $k$-space \footnote{\url{http://mridata.org/fullysampled/knees}} (referred as the knees dataset) to evaluate our model.  
This opensource images consists of 20 cases of  fully-sampled 3D Fast Spin Echo MR Images. 
We also chose randomly 10 slices in the middle of each case and further divided them into 2 sets: training and testing, 100 images each.  
Figure~\ref{fig:knee_zoom} depicts the representative reconstructions of knees test sets at the sampling rates 10\% and their 10$\times$ magnified error plots on the image magnitude (top 3 rows) and phase (bottom 3 rows). 
As can be seen, the proposed RefineGAN can fruitfully reconstruct the result which has less error compared to other methods. 
}

\rv{We also observed that RefineGAN consistently outperforms SingleGAN and ReconGAN for various sampling strategies. 
For example, Figure~\ref{fig:nrmse_various_2} visualizes the NRMSEs curves of the knees dataset using radial, cartesian, random and spiral sampling strategies (rate 20\%). 
We observed that the radial sampling pattern results in the best performance among all. 
%
Moreover, our method reduces sampling-specific effects, i.e., difference between sampling strategies become less severe. 
}



%% file: conclusion.tex


\label{sec:conclusion}
In this paper, we introduced a novel deep learning-based generative adversarial model for solving the Compressed Sensing MRI reconstruction problem. 
%
The proposed architecture, RefineGAN, which is inspired by the most recent advanced neural networks, such as U-net, Residual CNN, and GANs, is specifically designed to have a deeper generator network $G$ and is trained adversarially with the discriminator $D$ with cyclic data consistency loss to promote better interpolation of the given undersampled $k$-space data for accurate end-to-end MR image reconstruction. 
%
%
We demonstrated that RefineGAN outperforms the state-of-the-art CS-MRI methods in terms of running time and image quality, thus indicating its usefulness for time-critical clinical applications.

In the future, we plan to conduct an in-depth analysis of RefineGAN to better understand the architecture, as well as constructing incredibly deep multi-fold chains with the hope of further improving reconstruction accuracy based on its target-driven characteristic. 
Extending RefineGAN to handle dynamic MRI is an immediate next research direction. 
Developing a distributed version of RefineGAN for parallel training and deployment on a cluster system is another research direction we wish to explore. 